\documentclass[journal]{IEEEtran}

\usepackage[noadjust]{cite}
\usepackage{graphicx}

\ifCLASSOPTIONcompsoc
  \usepackage[caption=false,font=normalsize,labelfont=sf,textfont=sf]{subfig}
\else
  \usepackage[caption=false,font=footnotesize]{subfig}
\fi

\usepackage{amssymb}
\usepackage{amsmath}
\usepackage{txfonts}
\usepackage{lscape}
\usepackage{multicol}
\usepackage{comment}
\usepackage{color,soul}
\usepackage[shortlabels]{enumitem}
\usepackage{multirow}
\usepackage{array}
\newcolumntype{P}[1]{>{\centering\arraybackslash}p{#1}}

\begin{document}

\title{Blockchain-assisted Demonstration Cloning for Multi-Agent Deep Reinforcement Learning}

\author{Ahmed Alagha,
        Jamal Bentahar,
        Hadi Otrok,~\IEEEmembership{Senior Member, IEEE,}
        Shakti Singh, ~\IEEEmembership{Member, IEEE,}
        and Rabeb Mizouni % <-this % stops a space
\thanks{Ahmed Alagha and Jamal Bentahar are with the Concordia Institute for Information Systems Engineering, Concordia University, Montreal, QC H3G 1M8, Canada (e-mail: ahmed.alagha@mail.concordia.ca; jamal.bentahar@concordia.ca). Jamal Bentahar is also with the Department of Electrical Engineering and Computer Science, Khalifa University, Abu Dhabi, UAE (e-mail: jamal.bentahar@ku.ac.ae). Hadi Otrok, Shakti Singh, and Rabeb Mizouni are with the department of Electrical Engineering and Computer Science and the Center of Cyber Physical Systems (C2PS), Khalifa University, Abu Dhabi, UAE (emails: hadi.otrok@ku.ac.ae, shakti.singh@ku.ac.ae, rabeb.mizouni@ku.ac.ae).}}

\markboth{}%
{Alagha \MakeLowercase{\textit{et al.}}: Blockchain-assisted Demonstration Cloning for Multi-Agent Deep Reinforcement Learning}

\maketitle

\begin{abstract}
Multi-Agent Deep Reinforcement Learning (MDRL) is a promising research area in which agents learn complex behaviors in cooperative or competitive environments. However, MDRL comes with several challenges that hinder its usability, including sample efficiency, curse of dimensionality, and environment exploration. Recent works proposing Federated Reinforcement Learning (FRL) to tackle these issues suffer from problems related to model restrictions and maliciousness. Other proposals using reward shaping require considerable engineering and could lead to local optima. In this paper, we propose a novel Blockchain-assisted Multi-Expert Demonstration Cloning (MEDC) framework for MDRL. The proposed method utilizes expert demonstrations in guiding the learning of new MDRL agents, by suggesting exploration actions in the environment. A model sharing framework on Blockchain is designed to allow users to share their trained models, which can be allocated as expert models to requesting users to aid in training MDRL systems. A Consortium Blockchain is adopted to enable traceable and autonomous execution without the need for a single trusted entity. Smart Contracts are designed to manage users and models allocation, which are shared using IPFS. The proposed framework is tested on several applications, and is benchmarked against existing methods in FRL, Reward Shaping, and Imitation Learning-assisted RL. The results show the outperformance of the proposed framework in terms of learning speed and resiliency to faulty and malicious models.
\end{abstract}


\section{Introduction}
\label{Sec: Intro}
\IEEEPARstart{T}{he} recent advancements in Deep Learning (DL) and its integration in Deep Reinforcement Learning (DRL) have shown great promise in generating intelligent agents for applications such as robotics \cite{zhang2022reinforcement}, environmental monitoring \cite{shurrab2022iot, abououf2022self}, and video games \cite{silver2017mastering}. In DRL, an agent builds its own intelligence by learning a policy for decision-making based on its experience in the environment, guided by a numerical reward that the agent tries to maximize. DRL is further extrapolated into Multi-Agent Deep Reinforcement Learning (MDRL), in which multiple autonomous agents cooperate or compete to achieve their objectives in a common environment. MDRL has received increasing interest in the past few years, especially in fields related to video games \cite{berner2019dota}, autonomous driving \cite{antonio2022multi}, and robot swarms \cite{alagha2022target}.

Despite its great achievements, DRL comes with several challenges that hinder its usability. One challenge is sample efficiency, which refers to the difficulty of learning from insufficient interactions with the environment, due to the high cost of collecting such interactions. This problem has been recently tackled by several works using Federated Reinforcement Learning (FRL) \cite{fan2021fault, tianqing2021resource, nadiger2019federated}, in which multiple agents learn from their local experiences and periodically share their models to be aggregated in a global model, which is then shared to all agents. Another challenge associated with DRL, and amplified in MDRL, is the curse of dimensionality, which refers to the increasing size and complexity of the state and action spaces with the increasing number of agents \cite{nguyen2020deep}. This challenge is further amplified with the use of sparse rewards, 
which are rewards that are rarely present in the state space of the environment. One common sparse reward function is to assign a single reward value only when the task is successfully completed, such as delivering a checkmate in a game of chess or destroying the opponent team's Ancient in a game of Dota. This increases the difficulty of the learning, as the agents are rarely exposed to the reward, especially in the early stages of the learning where the agents act randomly in the environment to collect experiences. This has been addressed through reward shaping, in which specific reward functions are designed for each problem with the aim of exposing the agents to the reward frequently in the environment \cite{hu2020learning, sami2022graph, sami2022reward}. 

The existing methods in the literature suffer from several drawbacks when tackling the aforementioned challenges. While FRL helps in addressing the sample efficiency issue, faulty and inaccurate shared models could adversely impact the aggregated model (backdoor poisoning attacks), leading to severe difficulty in learning \cite{nguyen2021federated}. Additionally, in applications where the models are represented as deep neural networks (DNN), the architecture of the DNN might vary from one user to the other. In FRL, the global model is obtained by averaging the shared models, which cannot be simply achieved if the models are of different architectures, and would require additional steps (such as knowledge distillation) that increase the computational overhead. This constrains all the FRL nodes to use the same DNN architecture, which is inefficient as some nodes might have capabilities to train more complex architectures than other nodes. In knowledge distillation, a smaller model (student) is trained to mimic the behavior of a larger and more complex model (teacher). While this allows different DNN architectures to learn from each other, knowledge distillation algorithms usually train both the student and teacher models using the same data (i.e. same application), since they are only concerned with reducing the complexity of the model \cite{gou2021knowledge}. Therefore, applying the aforementioned methods in RL environments of inherently different dynamics leads to struggles in learning convergence \cite{qi2021federated}. This is mainly because these methods operate directly on the model weights by averaging models (FRL) or altering weights according to external models (student-teacher), without verifying if the source models are suitable for the current environment dynamics. On the other hand, while reward shaping helps in providing frequent feedback to the agents during training, shaped reward functions require considerable engineering and experimentation, and could frequently lead to unstable learning or convergence to local optima \cite{nair2018overcoming, vecerik2017leveraging}. 

In summary, existing works suffer from the following drawbacks:
\begin{enumerate}
    \item FRL methods can be severely impacted by faulty or malicious models, which affects the performance of the aggregated model.
    \item Typical FRL methods do not allow for different DNN architectures to be used in the same process.
    \item FRL and knowledge distillation methods cannot handle models from environments of different dynamics.
    \item Methods based on reward shaping require considerable engineering and could lead to local optima.
\end{enumerate}

To tackle the aforementioned issues, we propose a Blockchain-assisted Multi-Expert Demonstration Cloning (MEDC) for MDRL. Inspired by Imitation Learning (IL), MEDC is a novel method that uses expert demonstrations in guiding the learning of new MDRL agents. Rather than averaging MDRL models, the proposed method utilizes experiences from previously trained models (experts) to suggest actions for new MDRL agents to follow during the training. This enhances the quality of the new collected experiences and helps the agents get more exposed to sparse rewards. The proposed method utilizes the suggested actions from the expert models in guiding the learning, without the need of aggregating or averaging, which allows for models of different architectures to be utilized in the same learning process. Additionally, the proposed method is resilient to faulty and malicious expert models, since the new agents are still learning based on their own experiences and rewards, which would reflect bad actions if suggested by experts. To allow for model sharing across different users, a framework based on a Consortium Blockchain is proposed, in which users share or request expert models. Model sharing on centralized servers, as in the typical FRL systems, introduces several issues including single point-point-of-failure, the need of a single trusted server, and the vulnerability to security bottlenecks such as the modification of the information shared by FRL nodes \cite{nguyen2021federatedd}. Unlike centralized cloud computing, Blockchain helps in providing a decentralized, transparent, and autonomous platform for model sharing with no repudiation. A Consortium Blockchain, specifically, provides better privacy, scalability, and efficiency when compared to public Blockchains \cite{kadadha2022context}, and better allowance for collaboration and data sharing between entities when compared to private Blockchains, making them suitable for model sharing. The InterPlanetary File System (IPFS) is used to manage the storage of models, and Smart Contracts (SCs) on the Blockchain are designed to manage the model allocation across users, while considering the models' contexts and performances, as well as the users' reputations. In summary, the contributions of this work are as follows:
\begin{enumerate}
    \item The design of a novel multi-expert Demonstration Cloning (MEDC), a method that utilizes experiences from multiple trained models to guide the learning of new MDRL agents.
    \item A Consortium Blockchain-based framework for model sharing across MDRL users, with the assistance of IPFS for storage.
    \item A model allocation mechanism, implemented through smart contracts, which considers the models' quality and users' reputations when assigning models to users. 
\end{enumerate}

\begin{figure}[h]
    \centering
    \includegraphics[width=\columnwidth]{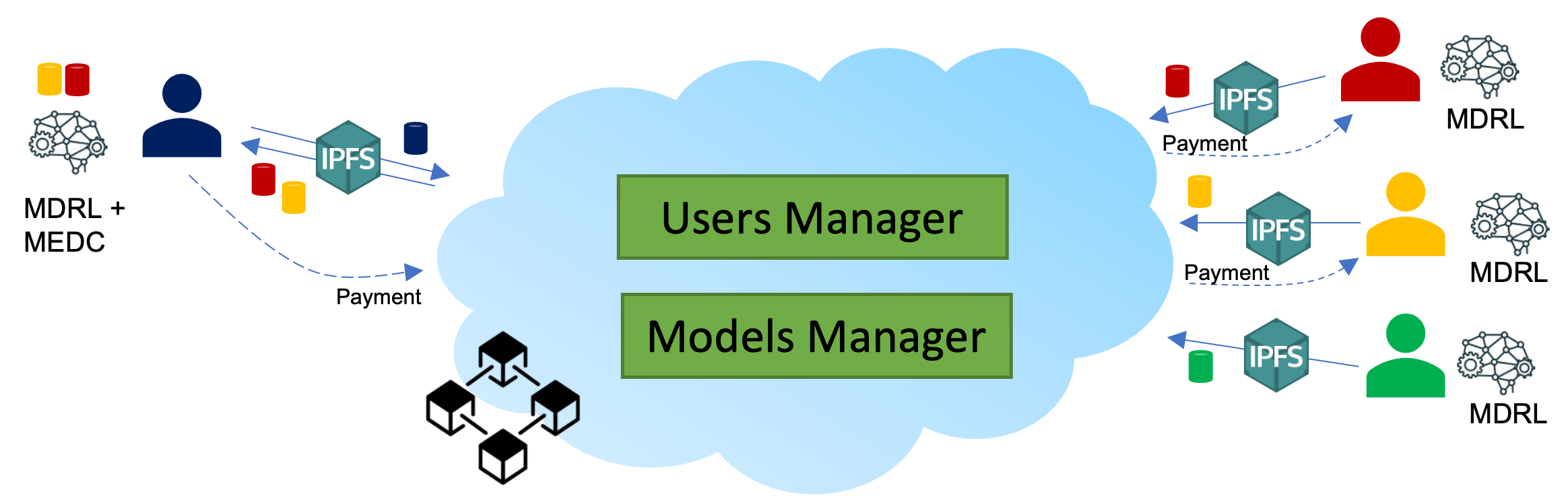}
    \caption{A general overview of the proposed framework.}
    \label{fig:GeneralOverview}
\end{figure}

Collectively, the entire proposed framework is responsible for managing model sharing on the Blockchain, and training through MEDC locally, and is generally described in Fig. \ref{fig:GeneralOverview}. Users locally train their MDRL models for their respective problems and share models via IPFS to the Blockchain, with the hope of receiving incentives. A user can request a set of models from the Blockchain for a certain pre-determined price. The requested models can be used locally with MEDC. The smart contracts on the Blockchain manage users' registrations and model submissions, in addition to allocating suitable models for users based on their requirements. The owners of the shared expert models are then paid accordingly, which incentivizes users to share their models.

The proposed methods are evaluated for the problem of target localization, where sensing agents are to be trained using MDRL to learn how to localize a radioactive target by progressively searching the area of interest. The environment of the problem could be of many different complexities, i.e. with varying number of agents or environment obstacles, which represents a suitable scenario for model and knowledge sharing across the different environments of the same problem. The adaptability of the proposed methods is tested on other applications, such as fleet coordination and maze cleaning, which are typical environments used to test MDRL algorithms. The proposed methods are tested and analyzed in terms of learning performance, showing scalability to different environments and resilience towards faulty models. The proposed methods are also benchmarks against works in FRL, Reward Shaping, and IL-assisted RL, showing dominance in terms of performance.

The remainder of this paper as follows: Section \ref{section: Related Work} reviews recent works in FRL, reward shaping, and IL-assisted RL. Section \ref{section: Proposed System} formulates the MDRL problem and describes the proposed Multi-Expert Demonstration Cloning (MEDC) method. Section \ref{Blockchain Framework} presents the proposed blockchain-based model sharing and allocation framework that complements MEDC. Experiments and results are presented and discussed in Section \ref{section: Sim & Eval}, and the paper is concluded in Section \ref{Conclusion}.

\section{Related Work}
\label{section: Related Work}

This section discusses existing literature tackling the MDRL challenges in data efficiency and reward sparsity, namely Federated Reinforcement Learning, Reward Shaping, and IL-assisted RL.
\subsection{Federated Reinforcement Learning}

Federated Reinforcement Learning (FRL) brings RL into the realm of FL. FRL aims to build a better policy from multiple RL agents without requiring them to share their raw experiences. Several works have adopted FRL, especially in the domain of MDRL, with the aim of increasing the sample efficiency of the training process, by aggregating models from different users trained on different experiences. In mobile edge computing, the authors in \cite{zhu2021federated} propose a multi-agent framework for data offloading. The problem of data allocation is formulated as a multi-agent Markov Decision Process (MDP), and a joint cooperation algorithm that combines the edge federated model with the multi-agent RL is proposed. Another work in \cite{wang2019edge} proposes a FRL framework for collaboration among edge nodes to exchange learning parameters, with the aim of better training and inference. FRL has also been combined with Blockchain, where the authors in \cite{yu2020deep} propose a framework that trains DRL models for computation offloading and resource allocation in 5G ultra-dense edge computing networks. The DRL models are trained in a distributed manner via a FL architecture, in which the communication is done securely over the Blockchain. In robotics applications, the authors in \cite{liu2019lifelong} propose a FRL architecture for cloud robotic technologies, in which a shared model on the cloud is upgraded with knowledge from different robots performing autonomous navigation. In autonomous driving, the authors in \cite{liang2022federated} propose an online FRL transfer process for real-time knowledge extraction, where agents take actions based on their own knowledge and the knowledge shared by others.

Despite the several advantages of FRL, it comes with several drawbacks. In a realistic scenario, if the models are trained in environments that inherit different dynamics, the learning convergence of the aggregated model could face issues \cite{qi2021federated}. Additionally, in most FRL frameworks, the global model is obtained by averaging the shared models, which requires all the models to have the same architecture in the case of neural networks. While this could be tackled by other additional steps, such as model compression (knowledge distillation), it introduces additional overhead and risk of losing information. Moreover, FRL is vulnerable to random failures or adversarial attacks, in which the shared models give harmful behavior that could affect the aggregated global model \cite{fan2021fault}. 

\subsection{Reward Shaping}

In DRL, a sparse reward is a case where the environment rarely produces a useful reward signal. Sparse rewards are the easiest and most common form of rewards, as the desired goals in most applications naturally induce a sparse reward, such as achieving a checkmate in chess. However, due to the complexity of DRL problems, especially in MDRL, sparse rewards induce difficulty in learning, especially during the exploration stage where the agents initially act randomly in the environment and barely collect rewards. Several works have introduced shaped reward functions, which distribute the reward over the course of the learning. One common form of shaped rewards are distance-based rewards, where agents get rewarded in each step of an episode if they get closer to achieving the goal. For example, in the problem of target localization \cite{liu2019double, alagha2022target, alagha2023multi}, agents get rewarded at each step if they get closer to the target, where measures like Euclidean Distance, Manhattan Distance, or Breadth-First Search are used in each step to compute the distance. In \cite{baker2019emergent}, the authors tackle the problem of Hide-and-Seek, where a vision-based reward is designed that rewards seekers in each step if they keep hiders within their sight, and rewards hiders in each step otherwise. Other proposed reward shaping methods alter the original reward with values generated from a shaping function. The authors in \cite{sami2022graph} propose a scheme for reward shaping based on Graph Convolutional Recurrent Networks to predict and produce reward shaping values. Another work in \cite{dong2020principled} proposes a reward shaping method based on Lyapunov stability theory, which tempts the RL process into maximal reward region by driving the reward to make the Lyapunov function.

Despite the fact that reward shaping is seen to speed up the learning in DRL and MDRL, designing shaped rewards requires considerable engineering, and could still lead to local optima \cite{nair2018overcoming, vecerik2017leveraging}. Additionally, many of the shaped rewards or reward shaping methods are computationally expensive, such as search methods or graph neural networks, which adds additional overhead to the learning.

\subsection{Imitation Learning-assisted RL}
\label{Subsec: IL RL Literature}
A recent alternative method to speed up the learning in RL, instead of reward shaping, is to combine RL with Imitation Learning (IL). In this approach, previously obtained experts (or expert demonstrations) are used partially to help in training new agents. Here, the demonstrations are used in a supervised learning method where the goal is to minimize the loss between the agent's actions and the expert's demonstrations. During IL, the agents do not collect rewards, and the learning is entirely based on the expert's demonstrations which act as labeled data in supervised learning. In \cite{nair2018overcoming, vecerik2017leveraging}, the authors propose methods that alternate between RL and IL for off-policy RL. In their case, the expert demonstrations are stored in a buffer, and for a portion of the RL period, the RL agent is trained with Behavioral Cloning (BC), where the aim is to exactly mimic the behavior of the expert with no reward feedback. The authors in \cite{sartoretti2019primal, damani2021primal} extrapolate these works into MDRL, where agents alternate between MDRL and behavioral cloning (IL) from an expert.

The above works prove efficient in speeding up the learning, under the assumption that an expert model that is proficient with the environment exists. However, if there is a slight variance in the expertise of such an expert (i.e. the expert is familiar with a similar environment that is not exactly the same as the agent's environment), this introduces difficulties in the learning convergence. This is because, in behavioral cloning, the agents have no way of determining whether the expert demonstrations are good or bad, and are just tasked to mimic the expert's behavior into their own policies.

The proposed Blockchain-assisted Multi-Expert Demonstration Cloning (MEDC) tackles the drawbacks in all the aforementioned works. First, the knowledge sharing process is only done based on the suggested actions by the expert models, which allows models to be of different architectures, unlike the case in FRL. Additionally, the use of simple sparse rewards is viable with MEDC, as expert models help in overcoming the exploration issues that accompany the use of sparse rewards. Finally, even though some of the actions are suggested by expert models, MEDC is resilient to faulty or malicious models, because agents in MEDC are still learning through their own reward values, which is not the case in most FRL and IL methods.

\section{MDRL with Multi-Expert Demonstration Cloning}
\label{section: Proposed System}

The proposed MEDC method utilizes previously trained expert models to guide the learning for the current MDRL system. This section first formulates the MDRL problem, then defines the MEDC method.

\subsection{MDRL Formulation and Policy Optimization}
\label{subsection: MDRL formulation}

MDRL is generally formulated as a Markov Game \cite{gronauer2021multi}, which generalizes the Markov Decision Process (MDP) from single-agent to multi-agent settings. In most MDRL problems, the agents cannot observe the full state of the environment, and hence the problem is modeled as a Partially Observable Markov Game (POMG). Each agent acts in the environment using a policy that translates the agent's observations into actions, with deep neural networks (DNNs) being the most common methods to represent policies \cite{nguyen2020deep}. The learning in a POMG unfolds over a finite sequence of steps, where at every step, each agent $i$ analyzes its observation $o_i$ and takes action $a_i$ based on a policy $\pi_i$ and receives a reward $r_i$. In MDRL, all agents at a certain step simultaneously select an action $\textbf{a} = (a_1, \dots, a_N)$ and receive a reward and observation $\textbf{o} = (o_1, \dots,o_N)$. The objective for each agent is to maximize the expected sum of rewards it receives during the game. 

Proximal Policy Optimization (PPO) \cite{schulman2017proximal} is used in this work as the RL algorithm due to its simplicity, low computational complexity, and balance between sample efficiency and wall-clock time. PPO uses an actor-critic structure, where an actor policy $\pi_\theta$, parametrized by $\theta$, is to be optimized with the objective of maximizing the cumulative reward throughout the episode, which is estimated through the value function (critic). PPO uses the current experiences, combined with the critique, and tries to take the biggest improvement step to update the current policy, without moving far from it. This tackles instability issues in policy gradient methods due to large policy updates. PPO optimizes a clipped policy surrogate, $L^{CLIP}(\theta)$, which is parameterized by $\theta$ and is given as

\begin{equation}
    L^{CLIP}(\theta) = \hat{\varmathbb{E}}_t \left[\text{min}(r_t(\theta)\hat{A}_t, \text{clip}(r_t(\theta), 1-\varepsilon, 1+\varepsilon)\hat{A}_t) \right],
    \label{eq. clipped surr}
\end{equation}
where $\hat{\varmathbb{E}}_t$ is the expectation operator at step $t$. The clip() function is responsible for clipping the value of $r_t(\theta)$ to be within the range of $[1-\varepsilon, 1+\varepsilon]$. Here, $\varepsilon$ is a clipping hyperparameter that determines the amount of clipping, which usually has a value of 0.2 \cite{schulman2017proximal}. In this function, 
\begin{equation}
r_t(\theta) = \frac{\pi_\theta(a_t|s_t)}{\pi_{\theta_\text{old}}(a_t|s_t)}
\end{equation} 
is the probability ratio of taking an action between the old policy $\pi_{\theta_\text{old}}$ and the current policy $\pi_\theta(a_t|s_t)$. $\hat{A}_t$ is the advantage function which estimates how good a certain action is in a given state. In this work, Generalized Advantage Estimate (GAE) \cite{schulman2015high} is used to estimate $\hat{A}_t$. After each $H$ timesteps of experience collection, where $H$ is the horizon length, the estimator of the advantage function is then given as

\begin{equation}
    \label{eq:advantage}
    \hat{A}_t^{\text{GAE}(\gamma, \lambda)} = \sum_{l=0}^{H} (\gamma \lambda)^{l} \delta_{t+l}, \quad \delta_{t+l} := r_{t+l} + \gamma V(s_{t+l+1}) - V(s_{t+l}),
\end{equation}
where $\delta_{t+l}$ is the Temporal Difference (TD) residual, $\gamma$,$\lambda \in [0,1]$ are discount factors, and $V(s_t)$ is the value predicted by the critic for the state in step $t$. The authors in \cite{schulman2017proximal} use additional parameters in Eq. \ref{eq. clipped surr} using entropy and value function error to ensure sufficient exploration and stable learning. 

The general learning process used in this work is shown in Algorithm 1. In this work, we use a Centralized-Learning \& Decentralized-Execution method, which is common in MDRL problems [29].
The learning unfolds over a set of episodes, where each episode represents a new instance of the problem. At the beginning of each episode, the environment resets to the initial state, and the agents collect the initial observations ($\textbf{o}^0 = (o_1^0, \dots,o_N^0)$). In episode step $j$, each agent $k$ uses their own observation ($o_k^j$) in their copy of the actor network to get a probability distribution over the possible actions ($P_k$), from which an action ($a_k^j$) is sampled. All agents then execute their joint actions ($\textit{\textbf{a}}^j$) in the environment to collect and store the next observations ($\textit{\textbf{o}}^{j+1}$), the reward value ($r^j$), and a termination flag ($f^j$) indicating if the episode is finished. The process is then repeated for the following episode steps, where the agents act based on new observations and collect and store new rewards and termination flags. If an episode terminates ($f^j$ equals 1), then a new episode is initiated with a reset. If the total number of training timesteps reaches the horizon ($H$), the collected experiences (observations, rewards, flags) are used to compute $\hat{A}$ and update the actor and critic networks using $L^{CLIP}(\theta)$. The process repeats until the specified total number of training timesteps is met.

\begin{table}[!ht]
%\label{BayesAlg}
\setlength{\tabcolsep}{3pt}

\centering
%\resizebox{\columnwidth}{!}{
\begin{tabular}{p{240pt}}
\hline
\textbf{Algorithm 1:} MDRL training with PPO\\ 
\hline
\textbf{Input:} Environment dynamics and actor/critic networks\\
1: \textbf{for} $i$ = 0,1,...,TotalSteps:\\ 
2: \hspace{0.25cm} $\textbf{o}^0$ = reset() \\
3: \hspace{0.25cm} \textbf{for} $j$ = 0,1,\dots,EpisodeLength:\\
4: \hspace{0.5cm} \textbf{for} $k$ = 1,2,\dots,TeamSize:\\
5: \hspace{0.75cm} $P_k$ = actor($o^j_k$) \color{blue}{\#probability distribution under agent $k$'s actor}\\
6: \hspace{0.75cm} $a^j_k$ = sample($P_k$) \color{blue}{\#sample an action from $P_k$}\\
7: \hspace{0.5cm} \textbf{end for}\\
8: \hspace{0.50cm} $\textbf{a}^j$ = [$a^j_1, a^j_2, ...$]\\
9: \hspace{0.50cm} $\textbf{o}^{j+1}, r^j, f^j$ = Step($\textbf{a}^j$) \color{blue}{\#step in the environment}\\
10: \hspace{0.50cm} Store $\textbf{o}^{j+1}$, $\textbf{a}^j$, and $r^j$\\
11: \hspace{0.35cm} \textbf{if} i \% $H$ == 0 \textbf{then}: \color{blue}{\#PPO update}\\
12: \hspace{0.6cm} Compute $\hat{A}$ using the collected experiences\\
13: \hspace{0.6cm} Use $L^{CLIP}(\theta)$ to update the actor and critic networks\\
14: \hspace{0.35cm} \textbf{end if}\\
15: \hspace{0.35cm} \textbf{if} $f^j == 1$ \textbf{then} break \color{blue}{\#if the episode is done}\\
16: \hspace{0.1cm} \textbf{end for}\\
17: \textbf{end for}\\
\hline

\end{tabular}
%}
\label{First Model training}
\end{table}

\subsection{Multi-Expert Demonstration Cloning (MEDC)}
The main idea of the proposed MEDC method is to use demonstrations from experts to guide the MDRL agents into collecting better experiences. In MDRL environments with sparse rewards, better experiences are defined as ones where the agents are more frequently exposed to the sparse reward. In this work, and for a given problem of interest, an expert is defined as a previously trained model with expertise in tackling the exact same environment. On the other hand, a semi-expert is one with expertise in tackling a similar environment that slightly differs in complexity. For example, in the problem of target localization with a team of 5 agents in an area that has 3 obstacles, an expert model is one that has been previously trained on the exact same environment, while a semi-expert model is one that has been trained on a target localization environment with only a single agent. The semi-expert is not fully capable of tackling the current environment (5 agents), but can still provide some guidance that would be better than acting randomly in the environment. One of the main issues faced in MDRL with sparse rewards is the rare occurrence of reward states, which is dominant in the early stages of the learning where the agents' policies produce randomized actions. The proposed method aims at utilizing demonstrations from experts and/or semi-experts to assist a new MDRL process by providing better experiences with more rewards during the learning stage. 

In this work, the proposed MEDC method utilizes the experiences from multiple experts (or semi-experts) to guide the learning of the new agents, while maintaining the ``learning from rewards" concept in RL. In the proposed method, the learning alternates between MDRL and MEDC. During MDRL, the agents take actions in the environment based on their policy networks, and collect rewards accordingly. During MEDC, the experts suggest actions to be followed by the MDRL agents to explore the environment. Here, even though the actions are suggested by the experts, the agents still learn from their own execution of those actions, i.e. based on the collected rewards. As a result, experiences with more frequent occurrence of reward states are expected. This method is resilient to faulty-, malicious-, and semi-experts, i.e. models that might occasionally suggest bad actions, as such actions could be identified through the collected rewards. This is unlike the methods using IL (discussed in Section \ref{Subsec: IL RL Literature}), in which bad- or semi-experts would significantly deteriorate the performance of the new agents who have no input in deciding whether the suggested actions are good or bad.

\begin{figure}[h]
    \centering
\includegraphics[width=0.9\columnwidth]{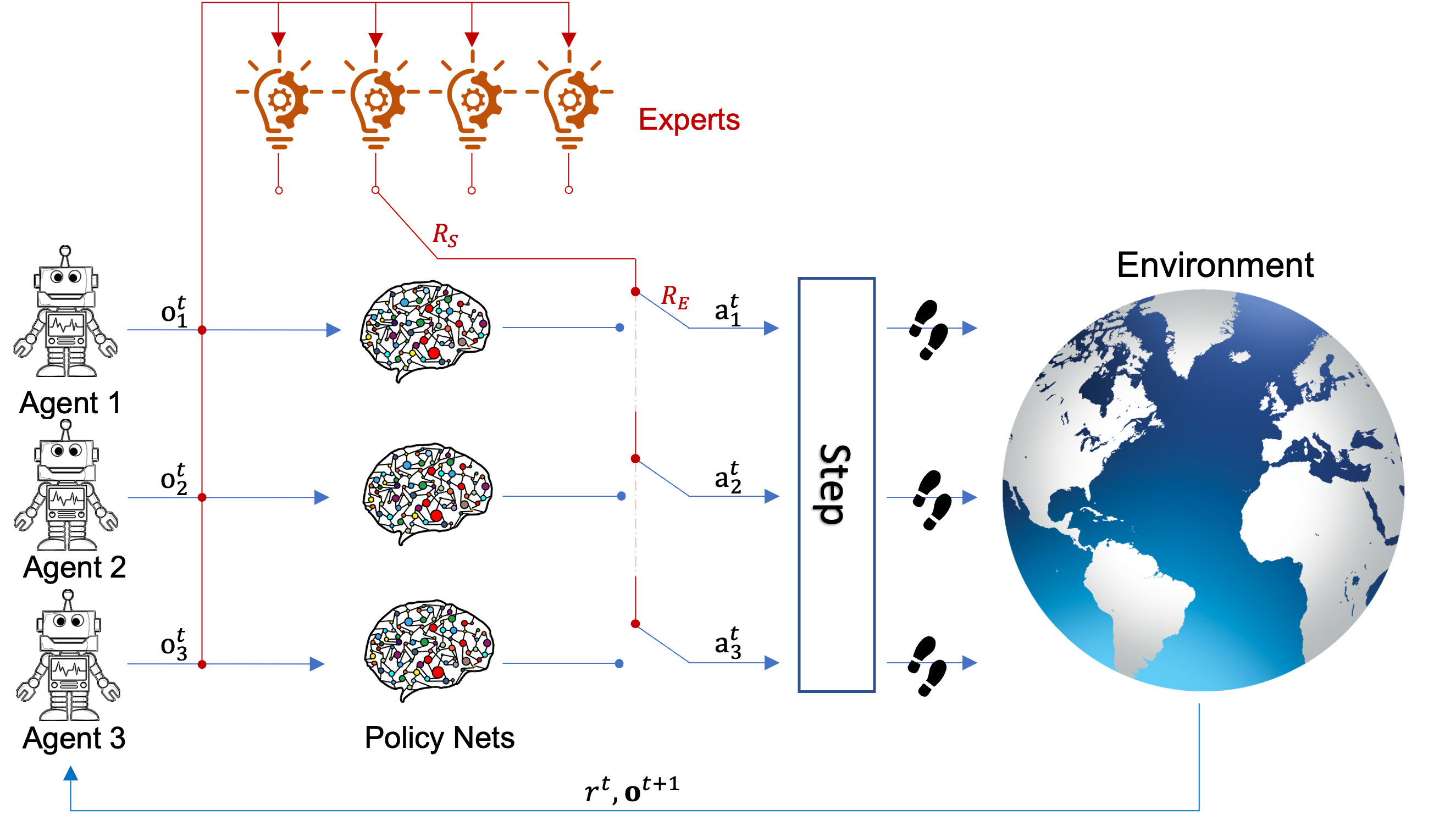}
    \caption{The proposed Multi-Expert Demonstration Cloning method.}
    \label{fig:MEDC}
\end{figure}

The learning process of MDRL with MEDC is described in Fig. \ref{fig:MEDC}. At the beginning of each episode, an expert probability $R_E$ determines whether the episode will follow MDRL or MEDC. During MDRL (i.e. the switch is closed), the typical process defined in Algorithm 1 is followed, where agents act based on their copies of the actor network. During MEDC, i.e. when the switch is on, one of the experts is selected, which takes the observations seen by the agents and suggests actions. The selection of the expert, out of the available experts, is done using a roulette wheel based on the attribute $R_S$, which is a common way used for selection \cite{lipowski2012roulette,alagha2022influence}. In this method, each candidate is associated with a probability which is proportional to $R_S$, where experts with higher $R_S$ have higher probability of being selected. After arranging the experts on a roulette wheel according to their probabilities, the wheel is spun to randomly select an expert. In this work, the attribute $R_S$ reflects a similarity measure between the environment of the expert and the current environment of interest. The method of computing $R_S$ depends heavily on the application. Some applications favor the number of agents as a measure of similarities, where environments with similar number of agents are favored, while other applications have problem-specific attributes related to the environment dynamics. In this work, we use a QoS (Quality of Service) metric to obtain the value of $R_S$, which is later discussed in Section \ref{Eval: target search}. Experts trained on environments/problems similar to the one of interest are more probable to be selected than ones trained on slightly different environments. Experts with lower similarities are still desired, as they introduce some variance in the experiences collected, which could be beneficial to the learning process. The actions suggested by the expert are then followed by the agents and the corresponding experiences are stored. This method is resilient to faulty-, malicious-, and semi-experts, i.e. models that might occasionally suggest bad actions, as such actions could be identified through the collected rewards.

Since PPO is an on-policy RL algorithm, the sampled actions taken by an agent should follow the agent's latest policy. If an agent follows actions that are not probable under their own policies, the learning would be unstable. To tackle this in the proposed MEDC method, the actions suggested by an expert are followed by the agents \textit{only if} they meet a probability condition $Q$. Given a probability distribution $P$ generated by the agent's policy over the actions, an action $a_e$ suggested by an expert is followed by the agent only if $P(a_e)>Q$. During MEDC, the expert ranks the possible actions, and the agent picks the best action that satisfies the $Q$ condition. This helps utilize demonstrations to collect better experiences, while not violating the on-policy learning. The value of $Q$ should be selected in a way that balances between the expert involvement and the on-policy learning. A very high value of $Q$ might limit the benefit of using an expert, as most of the suggested actions would not meet the threshold. On the other hand, a very low value of $Q$ results in most of the suggested actions being followed by the agents, even those that are far from the agents' own policy, which could lead to unstable learning when using the on-policy PPO. At later stages of the learning, the agents become more confident of their decisions, and the involvement of experts gets reduced since actions suggested by them will rarely meet the $Q$ threshold. This is because the probability distribution $P$ is concentrated in one action, and all other actions would have very low probabilities. 

Algorithm 2 describes the learning process of MDRL with MEDC. The process is similar to the one described in Algorithm 1, with the addition of MEDC (lines 3 and 7-14) that only affects the action selection process. Before the beginning of each episode, a randomly generated probability (ExpertCheck) determines if the episode follows MDRL or MEDC. If the value does not meet the Expert Rate (i.e. ExpertCheck $\geq R_E$), then the agents follow MDRL exactly as described previously in Algorithm 1. If the Expert Rate is met (i.e. ExpertCheck $< R_E$), the agents follow MEDC. Here, a roulette wheel selects which expert model to use according to their $R_S$ values. The agents' observations are fed into the chosen expert model, which returns the suggested expert actions (ExpActions). For each agent $k$, the expert actions that do not meet the threshold $Q$ under the agent's policy distribution $P_k$ are eliminated. The agents then follow the most valued expert actions in the environment to collect experiences. The process repeats throughout the episode, and the collected experiences are then used to update the actor and critic networks as described in Algorithm 1.

\begin{table}[!ht]
\setlength{\tabcolsep}{3pt}

\centering
\begin{tabular}{p{240pt}} 
\hline
\textbf{Algorithm 2:} Training MDRL with MEDC\\ 
\hline
\textbf{Input}: Environment dynamics, actor/critic networks, Expert Models, Similarity Measures ($R_S$), Expert Rate ($R_E$)\\
1: \textbf{for} $i$ = 0,1,...,TotalSteps:\\ 
2: \hspace{0.25cm} $\textbf{o}^0$ = reset()\\
3: \hspace{0.25cm} ExpertCheck = rand() \color{blue}{\#random probability}\\
4: \hspace{0.25cm} \textbf{for} $j$ = 0,1,\dots,EpisodeLength:\\
5: \hspace{0.5cm} \textbf{for} $k$ = 1,2,\dots,TeamSize:\\
6: \hspace{0.75cm} $P_k$ = actor($o^j_k$) \color{blue}{\#probability distribution under agent $k$'s actor}\\
7: \hspace{0.75cm} \textbf{if} ExpertCheck $< R_E$ \textbf{then}: \color{blue}{\#MEDC} \\
8: \hspace{1cm} ExpertModel = RouletteWheel(ExpertModels, $R_S$)\color{blue}\\
9: \hspace{1cm} ExpActions = ExpertModel($o^j_k$) \color{blue}{\#expert suggested actions}\\
10: \hspace{.85cm} \textbf{for} $x$ = 0,1,\dots,\textit{len}(ExpActions):\\
11: \hspace{1.1cm} if $P_k$[x]$< Q$: \\
12: \hspace{1.25cm} ExpActions[x] = 0 \color{blue}{\#exclude improbable actions under $P_k$}\\
13: \hspace{.85cm} \textbf{end for}\\
14: \hspace{0.9cm} $a^j_k$ = argmax(ExpActions) \color{blue}{\#take most probable action}\\
15: \hspace{0.6cm} \textbf{else:}\\
16: \hspace{0.9cm} $a^j_k$ = sample($P_k$) \color{blue}{\#sample an action from $P_k$}\\
17: \hspace{.6cm} \textbf{end if}\\
18: \hspace{0.35cm} \textbf{end for}\\
19: \hspace{0.35cm} $\textbf{a}^j$ = [$a^j_1, a^j_2, \dots$]\\
20: \hspace{0.35cm} $\textbf{o}^{j+1}, r^j, f^j$ = Step($\textbf{a}^j$) \\
21: \hspace{0.35cm} Store $\textbf{o}^{j+1}$, $\textbf{a}^j$, and $r^j$ \\
22: \hspace{0.35cm} \textbf{if} i \% $H$ == 0 \textbf{then}:  \color{blue}{\#PPO update}\\
23: \hspace{0.6cm} Compute $\hat{A}$ using the collected experiences \\
24: \hspace{0.6cm} Use $L^{CLIP}(\theta)$ to update the actor and critic networks\\
25: \hspace{0.35cm} \textbf{end if}\\
26: \hspace{0.35cm} \textbf{if} $f^j == 1$ \textbf{then} break  \color{blue}{\#if the episode is done}\\
27: \hspace{0.1cm} \textbf{end for}\\
28: \textbf{end for}\\
\hline
\end{tabular}
\label{Algorithm: MDRL with MEDC}
\end{table}

\section{Blockchain-based model sharing for Demonstration Cloning}
\label{Blockchain Framework}
This work proposes a Blockchain-based framework that complements the MEDC method. The framework is responsible for managing the users' registration, model submission, and appropriate model allocation to requesting users. A Consortium Blockchain is used due to its ability to offer increased privacy, shared control, efficiency, cost savings, and trust for multiple organizations or entities collaborating on a project or sharing data \cite{kadadha2022chain}. A Consortium Blockchain is operated by a group of entities, which introduces increased privacy and trust when compared to public Blockchains, and more collaboration allowance when compared to private Blockchains. This makes Consortium Blockchains suitable for data sharing across entities, and hence suitable for the purpose of model sharing in the proposed framework. 

\begin{figure}[h]
    \centering
    \includegraphics[width=\columnwidth]{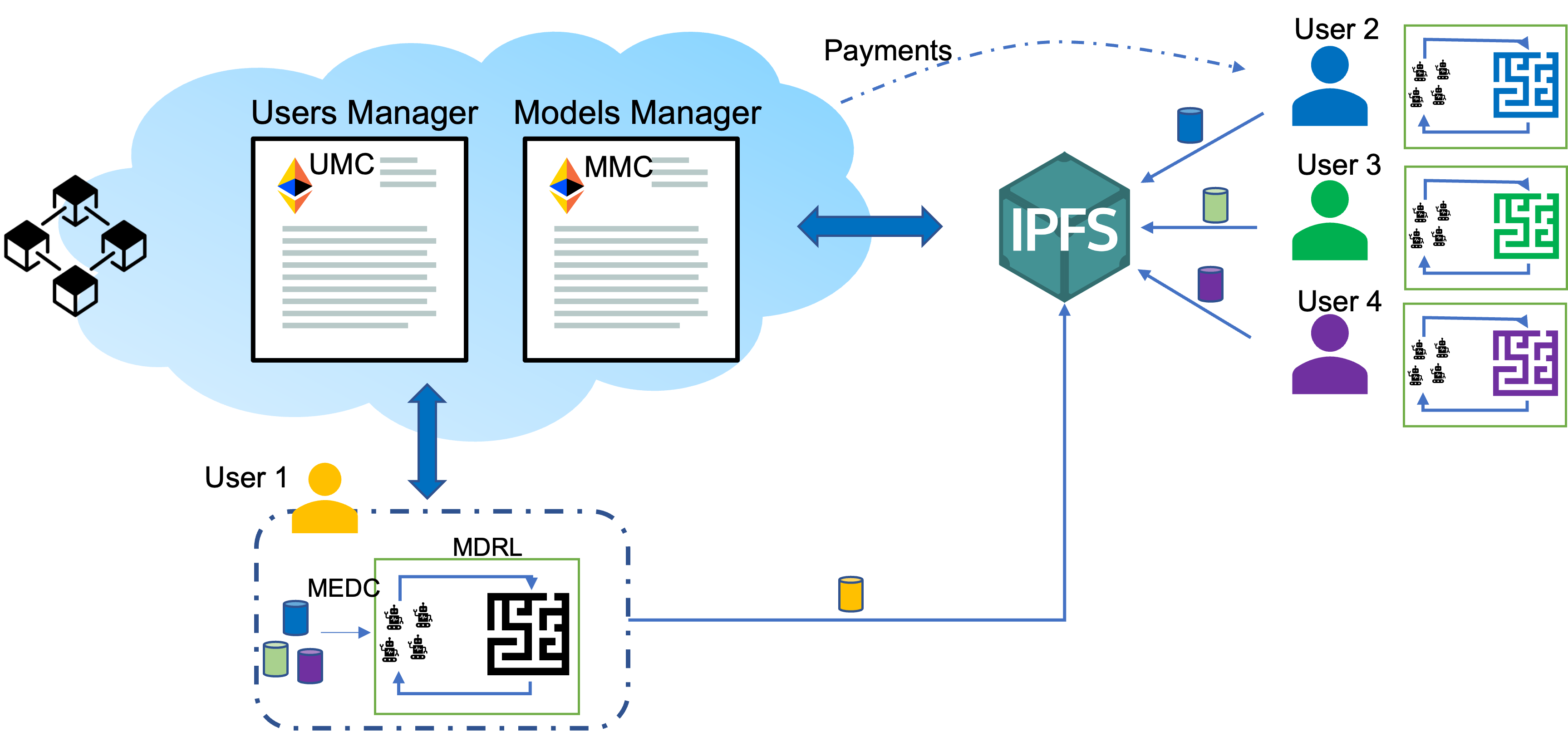}
    \caption{The proposed Blockchain-assisted model sharing framework for Demonstration Cloning.}
    \label{fig:framework}
\end{figure}

A high-level view of the proposed framework is shown in Fig. \ref{fig:framework}. The framework is built using two smart contracts: 1) Users Manager Contract (UMC) and 2) Models Manager Contract (MMC). The users interact with the UMC to register in the system and add their information. Users could also share their trained models by interacting with the MMC, with the hope of receiving incentives if the models are requested and used by other users. A user could share their trained model on IPFS, which returns a unique Content Identifier (CID) that can be used to access the file. The IPFS is a protocol designed to create a content-addressable Peer-to-Peer (P2P) decentralized file system \cite{benet2014ipfs}. Users share the CID with the MMC when submitting a trained model, along with information describing the problem and the environment details. In the proposed framework, the user indicates the problem of interest from a set of pre-defined problems. Additionally, the environment details are given as a tuple of pre-defined features that the user needs to specify. For example, in the field of target localization, the problem is defined as ``target localization'', while the environment details could be given as (Number of agents = 3, Number of targets = 1, Number of obstacles = 3). A user requests a set of models by interacting with the MMC and specifying the requirements. The MMC allocates suitable models after the requester submits the required payment, which is shared with the model owners. This incentivizes users to share their models with the hope of receiving payments. To further incentivize users to share efficient models, extra payments could be given if the shared model meets performance requirements. Alternatively, instead of paying fixed amounts for the shared models, model owners could have different valuations for their shared models based on their complexities and performances. Specific cost-efficient incentive mechanisms that motivate model owners to be truthful about their valuations are outside the scope of this work, but some potential works include auction-based incentive mechanisms \cite{wang2019optimization, wang2020worker}. For example, one possible way is to use variations of reverse auctions, where model requestors post their requirements on the blockchain, and model owners bid lower costs progressively. Another example is using a variation of second-price auctions, where model owners give valuations for their models and the lowest bid wins and gets picked by the requestor but gets paid the second lowest bid \cite{ehrhart2015auction}. 

\subsection{Smart Contract Implementation}
The details of the User Manager Contract (UMC) are shown in Table \ref{tab:UMC}. The \textit{User} data structure is designed to hold user information, including their Ethereum address and reputation. The \textit{Models Alloc. Count} reflects the number of times the user's models have been allocated to other users/requesters, while the \textit{Total Review} reflects the numerical sum of all the reviews submitted by requesters upon using the user's previously submitted models. Using these information, user $i$'s reputation is computed as

\begin{equation}
    \label{eq: Rep}
    Rep_i = \frac{\textit{Total Review}}{\textit{Models Alloc. Count}} .
\end{equation}

The UMC keeps users' information in the \textit{Users List} mapping, which maps a user's address to their \textit{User} object. The \textit{addUser()} function is responsible for registering users by creating a \textit{User} object and mapping it in the \textit{Users List}. The \textit{updateInfo()} function is invoked when one or more of the user's shared models are allocated to requesters, in order to update the user's information (\textit{Models Alloc. Count)}. The \textit{updateReputation()} function is responsible for updating the user's reputation as per Eq. \ref{eq: Rep}.

\begin{table}[ht]
    \caption{Users Manager Contract (UMC)}
    \centering
    \begin{tabular}{c|c|c}
    \hline
    \multicolumn{3}{c}{\textbf{Data Structure}}\\
    \hline
    \hline
    \multicolumn{3}{c}{\textbf{User}}\\
    \hline
    \textit{User Address (address)} & \multicolumn{2}{c}{\textit{Reputation (uint)}}\\
    \hline
    \textit{Models Alloc. Count (uint)} & \multicolumn{2}{c}{\textit{Total Review (uint)}}\\
    \hline
    \hline
    \multicolumn{3}{c}{\textbf{Variables}}\\
    \hline
    \multicolumn{3}{c}{\textit{Users List} (\textit{address} $\rightarrow$ \textit{User})}\\
    \hline
    \hline
    \textbf{Function} & \textbf{Parameters} & \textbf{Return}\\
    \hline
    \textit{addUser()} & \textit{address} & -\\
    \hline
    \textit{updateInfo()} & \textit{No. Models} & -\\
    \hline
    \textit{updateReputation()} & \textit{Review} & -\\
    \hline
    \end{tabular}
    \label{tab:UMC}
\end{table}

The details of the Models Manager Contract (MMC) are shown in Table \ref{tab:MMC}. The \textit{Model} data structure is designed to hold the information of a submitted MDRL model. This includes the \textit{Owner}'s Ethereum address, the \textit{CID} generated by IPFS, the \textit{Model Reputation}, the \textit{Allocation Count} of the model indicating the number of times the model has been allocated to requesters, the \textit{Total Model Review} based on previous requesters, a general \textit{Description} of the model indicating its architecture and input requirements, the \textit{Application} for which the model has been trained, and a tuple/array of \textit{Environment Details} giving values to each of the specific environment details. The MMC keeps the details of the shared models in a \textit{Models List} mapping, which maps an application to all the available models shared by users. The \textit{addModel()} function is responsible for registering the details of a newly submitted model. The \textit{allocateModel()} function is responsible for allocating models to a requester based on their requirements. The \textit{updateModelRep()} function is responsible for updating the model's reputation following a submitted review by a requester, which is computed for model $m$ as

\begin{equation}
    Rep_m = \frac{\textit{Total Model Review}}{\textit{Allocation Count}}.
\end{equation}

\begin{table}[ht]
    \caption{Models Manager Contract (MMC)}
    \centering
    \begin{tabular}{c|c|c}
    \hline
    \multicolumn{3}{c}{\textbf{Data Structure}}\\
    \hline
    \hline
    \multicolumn{3}{c}{\textbf{Model}}\\
    \hline
    \textit{Owner (address)} & \multicolumn{2}{c}{\textit{CID (string)}}\\
    \hline
    \textit{Model Reputation (uint)} & \multicolumn{2}{c}{\textit{Allocation Count (uint)}}\\
    \hline
    \textit{Total Model Review (uint)} & \multicolumn{2}{c}{\textit{Description (string)}}\\
    \hline
    \textit{Application (string)} & \multicolumn{2}{c}{\textit{Environment Details (uint[])}}\\
    \hline
    \hline
    \multicolumn{3}{c}{\textbf{Variables}}\\
    \hline
    \multicolumn{3}{c}{\textit{Models List (string $\rightarrow$ Model[])}}\\
    \hline
    \hline
    \textbf{Function} & \textbf{Parameters} & \textbf{Return}\\
    \hline
    \textit{addModel()} & \textit{Model Info} & -\\
    \hline
    \textit{allocateModels()} & \textit{Model Requirements} & \textit{Model[]}\\
    \hline
    \textit{updateModelRep()} & \textit{Review} & -\\
    \hline
    \end{tabular}
    \label{tab:MMC}
\end{table}

The \textit{allocateModel()} function employs a Greedy method to allocate models to the requester. A requester specifies a set of requirements including the application, the minimum model reputation, the minimum owner reputation, the desired environment details, and the number of models desired. The owner's reputation $Rep_i$ is important to consider as it reflects the owner's historical performance, which is essential especially when the current shared model is new and has not been reviewed yet. On the other hand, the model's reputation $Rep_m$ reflects the effectiveness of the shared model in guiding the learning, as per previous requesters. It is assumed that a requester pays the same amount of incentives for each requested model, and hence the number of models requested reflects the available budget. Models within the same application that do not meet the reputation requirements are filtered out. The rest of the models are ranked based on the following Quality of Service (QoS) metric:

\begin{equation}
    QoS_m = \frac{Rep_i \times Rep_m}{D_m},
    \label{eq:QoS}
\end{equation}
where $QoS_m$ is the QoS of model $m$, $Rep_i$ is the reputation of $m$'s owner, $Rep_m$ is the reputation of model $m$, and $D_m$ is a similarity measure between the environment in which model $m$ has been trained and the requester's environment. Given a set of $p$ environment attributes $E = [E_1, E_2, \dots, E_p]$, the value of $D_m$ is computed as 
\begin{equation}
    D_m = \sum_i^p w_i \times |E_i^m - E_i^r| \; , \; \; \; \; \sum_i^p w_i = 1  ,
    \label{eq: Dm}
\end{equation}
where $E^m$ is the set of environment attributes associated with the model, $E^r$ is the set of attributes associated with the requester's environment, and $w_i$ are weighting parameters. The weights $w_i$ in Eq. \ref{eq: Dm} reflect the importance of each attribute. Higher values give more significance to the difference between the environment of the model and that of the requester, which lowers the QoS if that difference increases. All environment attributes are given equal weights by default, unless specified otherwise by the requester in the \textit{Model Requirements} when invoking the \textit{allocateModel()} function.

\subsection{Framework Time Sequence}
\label{subsection: TimeSequence}

\begin{figure*}[h]
    \centering
    \includegraphics[width=0.98\textwidth]{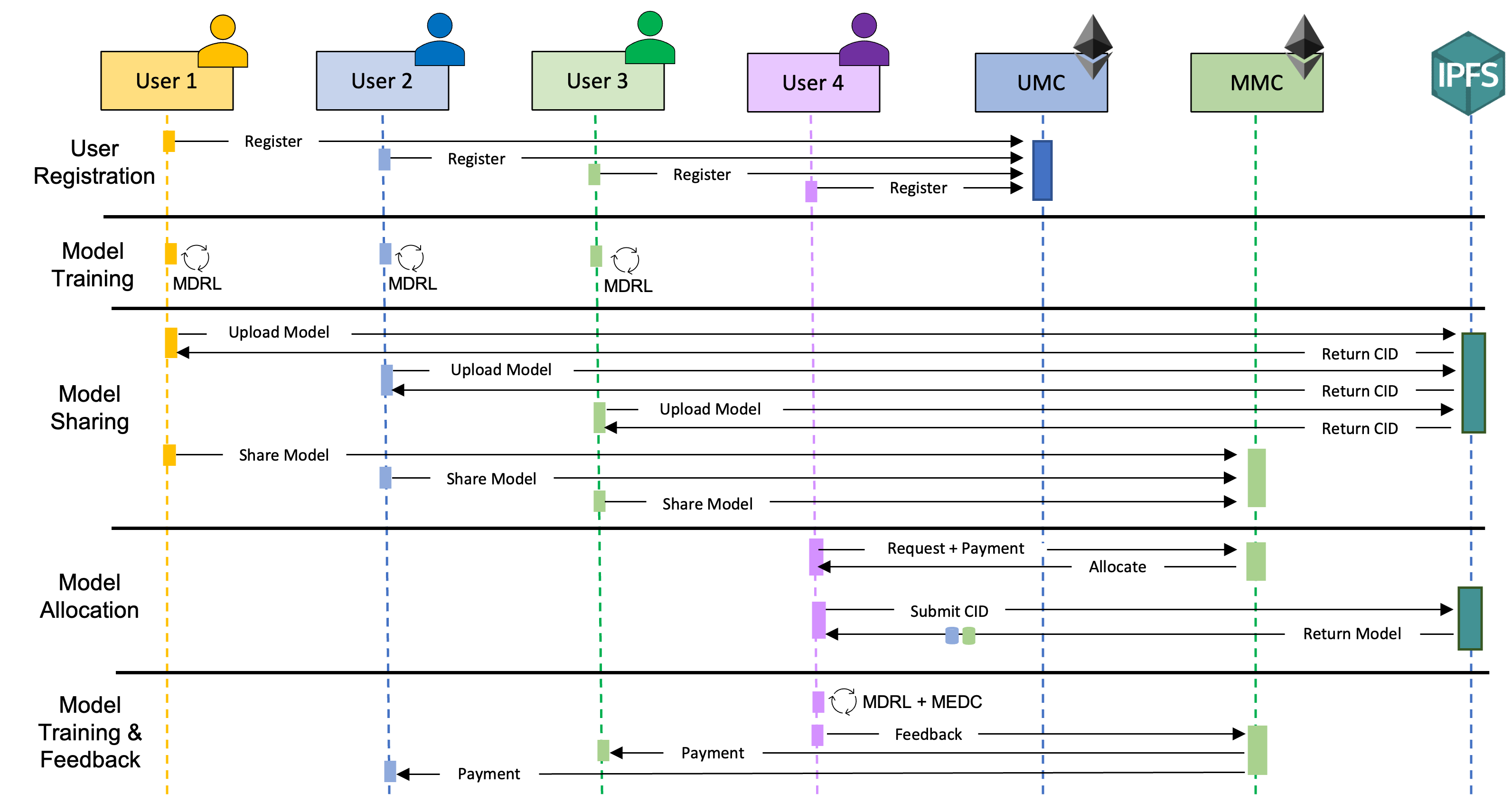}
    \caption{The interactions between the users and smart contracts as part of the proposed framework.}
    \label{fig:timesequence}
\end{figure*}

Fig. \ref{fig:timesequence} shows a time sequence diagram for a scenario under the proposed Blockchain-based framework for MEDC. It presents the interactions between the users and the smart contracts constituting the framework, which are given as follows:
\begin{itemize}
    \item \textbf{User Registration}: Users register to the UMC by invoking the \textit{addUser()} function. The function creates a \textit{User} object and appends it to the \textit{User List}. Each user is assumed to have a single Ethereum address linked to their account. The reputations are initialized with a value of 0.5, while \textit{Models Alloc. Count} and \textit{Total Review} are initialized with a value of 0. 
    \item \textbf{Model Training (MDRL)}: Users locally train MDRL models for their respective problems/applications. This step could occur at any time (before or after registration). Previously trained models could be stored by the users and shared later.
    \item \textbf{Model Sharing}: Users who wish to share their models upload the model files to IPFS, which returns a CID for each submitted model. The users then submit model details to the MMC by invoking the \textit{addModel()} function and passing the required information. The MMC creates a \textit{Model} object and appends it to the \textit{Models List}.
    \item \textbf{Model Allocation}: A user who wishes to request a set of expert models communicates with the MMC by invoking the \textit{allocateModel()} function and passing the requirements. The MMC runs the allocation mechanism and returns the models and their details to the user. The user accesses the models files on IPFS using the provided CIDs.
    \item \textbf{Model Training (MDRL + MEDC) and feedback}: The user utilizes the proposed MEDC with the allocated expert models to help in training MDRL agents for the problem of interest. The user then submits feedback reviewing the obtained models, which are used to update the models' and users' reputations.
    \item \textbf{Payments}: The owner of each requested model is paid a pre-determined amount in return for the shared model.
\end{itemize}

\section{Simulation and Evaluation}
\label{section: Sim & Eval}

This section presents and discusses different experiments conducted to validate the proposed methods. The experiments are first conducted in a custom environment for a task of Target Localization, which is a complex multi-agent problem requiring agents to cooperate in finding the target location. Subsequently, the adaptability of the proposed method is tested across two additional typical Multi-Agent applications, which are Fleet Coordination for Autonomous Vehicles \cite{xidias2016path, punma2018autonomous} and Multi-Agent Maze Cleaning \cite{jiang2021multi}. All the simulations have been conducted using an Intel E5-2650 v4 Broadwell workstation equipped with 128 GB RAM, 800 GB SSD, and NVIDIA P100 Pascal GPU (16 GB HBM2 memory).

\subsection{Application Environment: Target Localization}
\label{Eval: target search}
Target Localization is a multi-agent problem in which the location of a certain target is to be identified using sensory data reading collected by multiple mobile sensing agents \cite{alagha2019data, SHURRAB2023100867, hussain2023predictive}. Such a problem is of interest in applications including radiation monitoring \cite{alagha2020rfls, alagha2021sdrs}, search and rescue missions \cite{yuan2022uav}, and last-mile delivery \cite{abualola2023matching}. In such a problem, the sensing agents (UAVs or robots) need to cooperate to minimize the search time and the consumption of resources.

We formulate the target localization problem similar to \cite{liu2019double, alagha2022target}. The problem is formulated as a POMG, since the target location is unknown. The environment of the problem consists of an unknown target location, sensing agents, and walls/obstacles which complicate the mobility of the agents and attenuate the data readings collected. The agents act based on their observations, which consist of data readings and information about the environment (obstacles) and the locations of the team members. Using these observations, the agents need to explore (through moving actions) the environment in a way that minimized the search time. It is assumed that the agents are capable of communicating with each other, and hence they have information about the locations and readings collected by other agents. At the beginning of each training episode, the environment is initialized with a randomly placed target, randomly distributed obstacles, and randomly placed agents. An episode of target localization unfolds over a finite sequence of steps. At every step, each agent $i$ analyzes its observation $o_i \in \mathcal{O}_i$ and takes a moving action $a_i\in\mathcal{A}_i$ based on a policy $\pi_i : \mathcal{O}_i\times\mathcal{A}_i \rightarrow [0,1]$ and receives a reward $r_i$. We use a sparse reward function, where the agents receive a large reward only when the target is localized. During the training (which is conducted in simulations), the experiences and rewards collected are used in PPO, as discussed in Section \ref{subsection: MDRL formulation}, to update the policies of the agents. Once the training is over, the learnt policies could be deployed on real-world sensing agents. Table \ref{Table: Hyperparameters} shows the training hyperparameters used with PPO and MEDC. Here, the expert similarity measure $R_S$ that is used to determine the usage rate of each expert model is proportional to the QoS of that model.

\begin{table}[ht]
\caption{Hyperparameters used for PPO and MEDC.}
\setlength{\tabcolsep}{0pt}
\begin{center}
\begin{tabular}{|P{0.55\columnwidth}|P{0.25\columnwidth}|}
\hline
PPO Hyperparameters & Value\\
\hline
Learning rate & $3\times 10^{-4}$\\
PPO clipping parameter $\varepsilon$ & $0.2$\\
Entropy coefficient $c_2$ & $0.01$\\
Discount factor $\gamma$ & $0.99$\\
Discount factor $\lambda$ & $0.95$\\
Timesteps per update (Horizon $H$) & $4000$\\
Number of epochs per update & $20$\\
\hline
MEDC Hyperparameters & Value\\
\hline
Expert Rate $R_E$ & $0.1$\\
Expert Similarity $R_S$ & $\frac{QoS_i}{\sum{QoS_j}}$\\
Action Probability Threshold $Q$ & $0.05$\\
%\hline
\hline
\end{tabular}

\end{center}
\label{Table: Hyperparameters}
\end{table}

\begin{figure}[h]
     \centering
     \subfloat[Area of interest\label{Fig: AoI}]{
     \includegraphics[width=0.3\columnwidth]{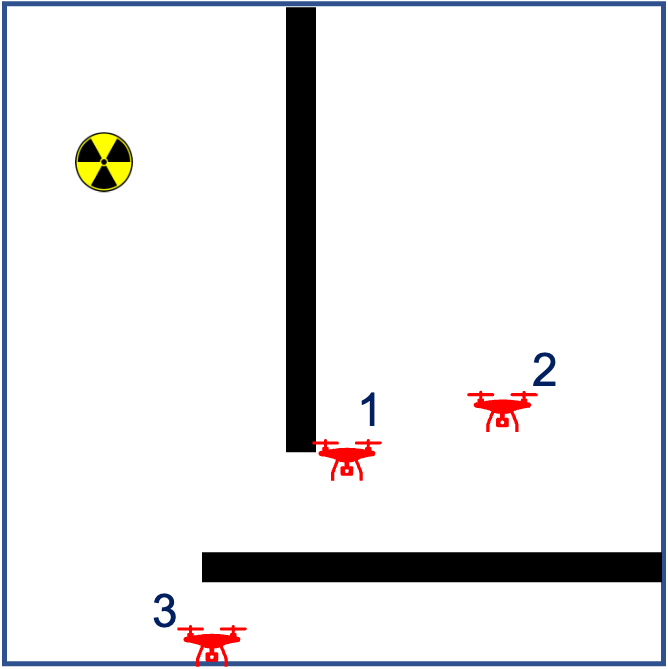}}
     \hfill
     \subfloat[Location Map\label{Fig: LocationMap}]{
     \includegraphics[width=0.3\columnwidth]{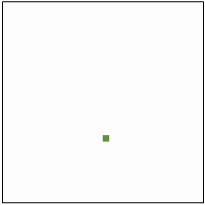}}
     \hfill     
     \subfloat[Team Locations Map\label{Fig: TeamLocationsMap}]{
     \includegraphics[width=0.3\columnwidth]{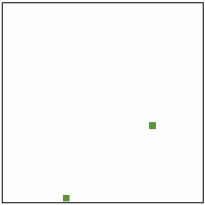}}
     \hfill     
     \subfloat[Readings Map\label{Fig: ReadingsMap}]{
     \includegraphics[width=0.3\columnwidth]{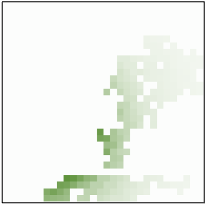}}
     \hfill     
     \subfloat[Visit Counts\label{Fig: VisitCountsMap}]{
     \includegraphics[width=0.3\columnwidth]{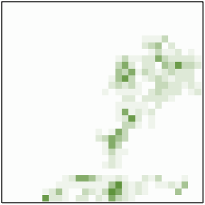}}
     \hfill     
     \subfloat[Walls Map\label{Fig: WallsMap}]{
     \includegraphics[width=0.3\columnwidth]{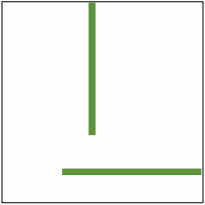}}
     \hfill     
        \caption{The set of observations (b)-(f) given the snapshot of the environment shown in (a) from the perspective of agent 1.}
        \label{Fig: Observations}
\end{figure}

An example of a target localization scenario is represented in Fig. \ref{Fig: AoI}, showing agents locations and obstacles distribution. At each step, and given a set of observations, each agent takes a moving action vertically, horizontally, or diagonally (in a 2D space) in a certain direction, with a pre-defined speed. As in \cite{liu2019double, alagha2022target}, the observations of each agent are modeled as 2D maps representing the area of interest, and fed to a Convolutional Neural Network (CNN) that acts as the policy producing actions. Figs. \ref{Fig: LocationMap}-\ref{Fig: WallsMap} show an example of the collected observations. The scenario here shows three sensing agents aiming to find a radioactive source in an area of size 5km$\times$5km. The Location Map defines the agent's own location, the Team Locations Map shows the distribution of other agents, the Readings Map encapsulates the collected readings throughout the area, the Visit Counts Map keeps track of the frequency different areas are visited at, and the Walls Map shows the distribution of walls/obstacles in the environment. We use a similar architecture for the CNN policy network as used in \cite{alagha2022target}, which is based on the LeNet-5 architecture \cite{lecun2015lenet}. The observations are fed to the architecture as a stack of five 2D maps. 

Target localization is chosen as a real-world problem to test the proposed methods due to the different possible variations the environment, which would be a good fit to test the proposed MEDC method and compare it with FRL. The environment could vary in terms of the number of agents or the number of walls, where each combination of agents and walls gives a different learning problem with its own complexity. In the following experiments, we first conduct analysis on the proposed MEDC and its effectiveness in achieving good learning, in addition to its resilience to faulty or malicious models, then compare it against existing benchmarks.

\subsection{Performance of MEDC}
\label{subsection: performance MEDC}
This section analyzes the performance of the proposed MEDC method and its effectiveness in guiding the learning for new agents in an MDRL system. During training, a localization episode has a maximum length of 100 steps, and the total number of training steps is given as $2\times10^7$. An episode terminates when the target is found, or when the maximum episode length is reached. After every 40,000 training steps, the average results of 4,000 testing steps, where the agents act greedily based on the latest policy update, are recorded and plotted. In this work, the target localization problem is characterized by two features: the number of agents and the number of walls. We report the results in terms of episodic length throughout the learning, as it reflects the time needed by the agents to localize the target, which is the main aim of target localization. We use the notation $AyWz$ to denote an environment with $y$ agents and $z$ walls.

\begin{figure}[h]
     \centering
     \subfloat[Number of walls = 2\label{MEDC_A3W2}]{
     \includegraphics[width=0.48\columnwidth]{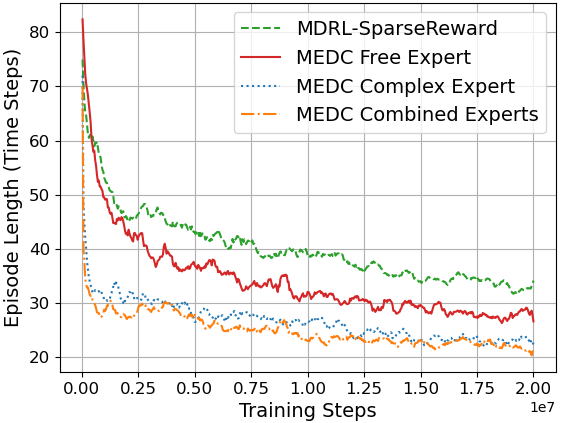}}
     \hfill
     \subfloat[Number of walls = 3\label{MEDC_A3W3}]{
     \includegraphics[width=0.48\columnwidth]{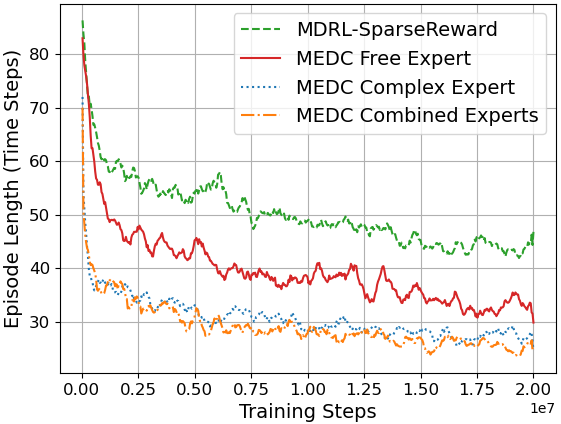}}
     \hfill
        \caption{The episodic length throughout the learning for an environment of 3 agents and (a) 2 walls and (b) 3 walls.}
        \label{Results: MEDC}
\end{figure}

To show the effectiveness of MEDC in enhancing the learning and tackling issues with sparse rewards, Fig. \ref{Results: MEDC} compares variations of an environment trained using MDRL and MEDC with one that is trained using MDRL with sparse rewards. Here, the problem of interest is one with 3 agents aiming to localize the target in an environment of 2 walls (A3W2) or 3 walls (A3W3). Three types of experts are used to guide the learning: a ``Free Expert'' is a model previously trained to tackle an environment with a single agent and no walls (A1W0), a ``Complex Expert'' is a model previously trained to tackle an environment with a single agent and 3 walls (A1W3), and ``Combined Experts'' refers to using both aforementioned experts at the same time in MEDC to guide the learning. It is evident that the used experts come from environment that are not exactly the same as the current environments of interest. Here, a better learning is indicated by a smaller episode length, as it reflects faster localization. As shown in Fig. \ref{Results: MEDC}, using MEDC with experts from similar environments to guide the MDRL agents helps in achieving better and faster learning, when compared to training MDRL independently using a sparse reward. Even though both experts are not familiar with a multi-agent environment, and can only tackle a single agent environment (hence they are not familiar with cooperation), they are partially beneficial in guiding the new agents into collecting better experiences, especially in the initial stages of the learning. Additionally, it can be seen that using the complex expert (A1W3) gives better results than using the free expert (A1W0), because it is trained in an environment closer to the current environment of interest. This reflects the importance of considering the QoS (Eq. \ref{eq:QoS}) when allocating expert models. In this scenario, and assuming that $Rep_m$ and $Rep_i$ are constant across both experts, the complex expert would have a higher QoS than the free expert since it has smaller $D_m$ values (assuming the features have the same importance).

\begin{figure}[h]
     \centering
     \includegraphics[width=0.55\columnwidth]{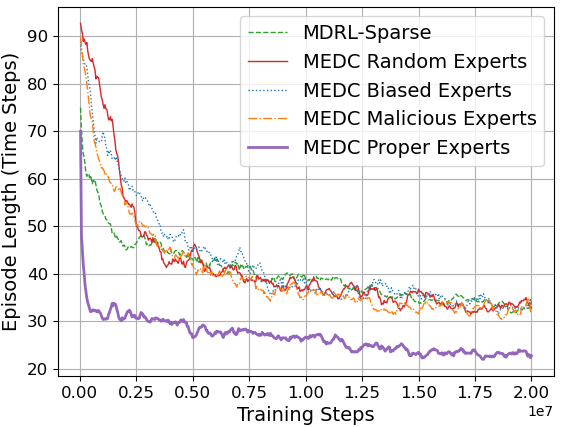}
     \hfill
        \caption{The episodic length throughout the learning for an environment of 3 agents and 2 walls, while using different faulty and malicious experts.}
        \label{Results: MEDC Resliency}
\end{figure}

To study the resiliency of the proposed MEDC method, the following experiment uses faulty and malicious experts. Fig. \ref{Results: MEDC Resliency} shows the learning performance in an environment of 3 agents and 2 walls, while using the following faulty experts:
\begin{itemize}
    \item Random Expert: a model that suggests random actions in the environment.
    \item Biased Expert: a model that always suggests the same action regardless of the collected observations.
    \item Malicious Expert: a model that is aware of the right action, but suggests the opposite action (opposite direction).
\end{itemize}

For each MEDC scenario, 5 of the mentioned expert types are used. For example, in ``MEDC Random Experts'', the agents are being trained using MDRL and MEDC with 5 Random Experts. ``MEDC Proper Experts'' is a scenario where 5 good experts are used from different environments, while ``MDRL-Sparse'' shows a scenario of training the MDRL agents independently using a sparse reward. As seen in the figure, the proposed MEDC is resilient to faulty and malicious expert models. At worst, the performance of MEDC drops to match that of ``MDRL-Sparse''. This means that, the use of experts in MEDC could either help the learning in MDRL or cause no harm. This is mainly because the experts are used for a small portion of the learning ($R_E = 0.1$). Additionally, the main purpose of the experts is to expose the agents to the sparse reward more frequently. If this does not occur, then the agents are still learning based on their own experience, even if these experiences are induced by the experts, which explains why the performance at worse is similar to the MDRL with sparse reward and no experts.

\subsection{MEDC vs Benchmarks}
\label{MEDC vs Benchmarks}
This section discusses and highlights the main advantages of MEDC when compared frameworks in FRL, Reward Shaping (RS), and IL-assisted RL. The benchmarks are summarized as follows: 

\begin{itemize}
    \item In FRL, models from different users are averaged frequently in a global model, which is shared back to them. We benchmark with the works in \cite{liu2019lifelong, liang2022federated}, where FRL is used to combine DRL models across different users.
    \item In Reward Shaping (RS), a shaped reward function is used to guide the learning. Here, we use a distance-based reward function, where the agents receive a positive reward in each step if they move closer to the target during the training, which is similar to the works in \cite{alagha2022target,liu2019double}. In each episode, the distances are computed using Breadth-First Search (BFS). BFS is used due to the existence of walls, which makes simpler distance measures (such as Euclidean distance) inapplicable.
    \item In IL-assisted RL, the learning alternates between RL and IL, similar to the works in \cite{damani2021primal, sartoretti2019primal}. For fairness, We extrapolate this work to use the same structure of MEDC with multiple experts. The main difference is, when IL is switched on, the MDRL agents use Behavioral Cloning (BC) to mimic the behavior of the expert without any reward input.
\end{itemize}

\begin{figure}[h]
     \centering
     \subfloat[Environment: A3W2\label{Benchmarks_A3W2}]{
     \includegraphics[width=0.48\columnwidth]{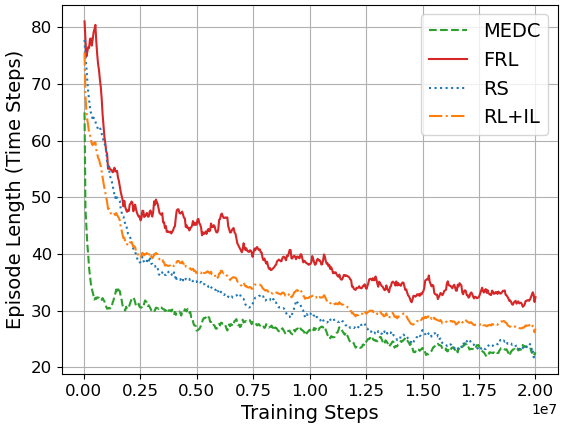}}
     \hfill
     \subfloat[Environment: A3W3\label{Benchmarks_A3W3}]{
     \includegraphics[width=0.48\columnwidth]{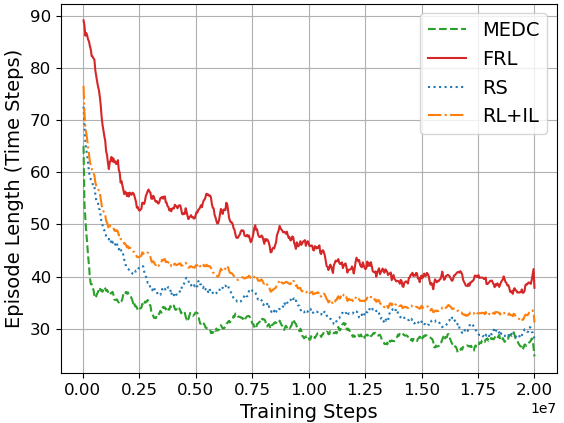}}
     \hfill
        \caption{The episodic length throughout the learning for an environment of 3 agents and (a) 2 walls or (b) 3 walls, while comparing MEDC with the benchmarks.}
        \label{Results: Benchmarks}
\end{figure}

Figure \ref{Results: Benchmarks} compares Blockchain-assisted MEDC with the 3 benchmarks. For MEDC and FRL, the learning involves 8 users of varying environments. The 8 users are training on different environments, given as A1W0, A3W0, A2W1, A2W2, A2W3, A3W1, A3W2, A3W3. For FRL, during the training, the models are shared and a global model is returned to each user to carry on the training. In MEDC, the first 6 users share their models to the Blockchain, which are then used by A3W2 and A3W3 in MEDC to guide the learning. For IL-assisted RL, the models of the first 6 users are used as experts, from which the MDRL agents in A3W2 and A3W3 choose to imitate. We show the training results from the perspective of the A3W2 and A3W3 users. As seen in the figure, the performance of the models trained with MEDC is significantly better than that of FRL. As discussed in Section \ref{section: Related Work}, if the models are trained in environments that inherit different dynamics, the learning convergence of the aggregated model could face issues \cite{qi2021federated}, which is seen in the obtained results. This is unlike MEDC, in which the expert models only suggest actions, and the MDRL agents still learn based on their experience. Similarly, MEDC outperforms IL-assisted RL, whose performance is negatively affected by experts being from different variations of the environment. This is because the agents in IL-assisted RL blindly clone the behavior of the expert into their models. Additionally, the performance of MEDC when compared to the model trained on a shaped reward shows slight outperformance. Even though RS gives a similar performance towards the end of the training, it is worth noting that the model trained with the BFS shaped reward required \textit{double} the wall time to train for $2\times10^7$ steps. This due to the computational power wasted on executing BFS in each episode, which reflects the complexity of most shaped rewards. MEDC, on the other hand, is able to achieve similar, if not better, results, with a simple sparse reward.

\begin{figure}[h]
     \centering
     \includegraphics[width=0.55\columnwidth]{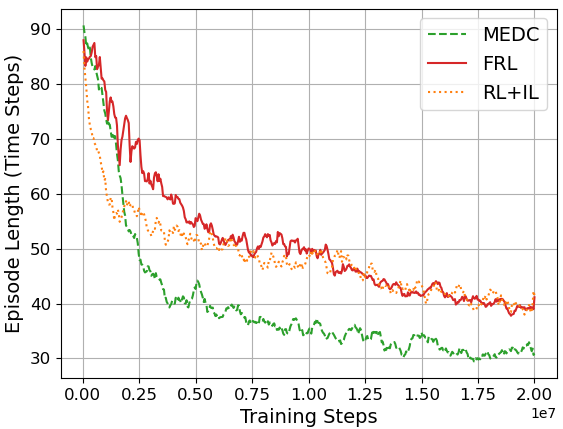}
     \hfill
        \caption{The episodic length throughout the learning for an environment of 3 agents and 2 walls, while using faulty and malicious experts.}
        \label{Results: MEDC vs FRL}
\end{figure}

To evaluate the resiliency of MEDC in comparison with FRL and IL-assisted RL, Fig. \ref{Results: MEDC vs FRL} compares the methods in the same scenario of 8 users, out of which 3 random users have the faulty experts discussed in section \ref{subsection: performance MEDC}. It is evident that the proposed MEDC method is more resilient and is able to maintain faster and better learning than FRL and IL-assisted RL, which are negatively affected by the bad actions suggested by the faulty and malicious experts.

\subsection{Adaptability to Other Applications}
This section analyzes the adaptability of the MEDC method to the following applications:
\begin{itemize}
    \item Fleet Coordination for Autonomous Vehicles \cite{xidias2016path, punma2018autonomous}: in this problem, a team of autonomous vehicles is tasked with picking up and dropping costumers at specific locations. Each vehicle has a limit in terms of the number of costumers it can accommodate simultaneously. At the beginning of each episode, all agents (vehicles) and costumers are placed randomly in the environment, with each costumer having a desired destination. The vehicles are tasked with cooperation and coordination, such that the time needed to drop all costumers at their destinations is minimized. This requires the agents to learn how to divide the costumers according to their locations. The agents receive a small reward for picking up a costumer, a large reward for dropping all costumers correctly, and a small negative reward for each time step to account for time cost. The agents observe their locations, the costumers' locations, and the desired destinations in a 2D format similar to Fig. \ref{Fig: Observations}. 
    \item Multi-Agent Maze Cleaning \cite{jiang2021multi}: in this problem, the agents are placed in a maze with the task of cleaning the maze as fast as possible. At the beginning of the episode, a maze is randomly generated, where the maze is considered entirely dirty initially, and each spot covered by a cleaning agent is considered to have been cleaned. The agents are required to coordinate and distribute in the maze in a way that minimizes the time needed for the maze to be fully cleaned. Each agent observes its own location as well as other agents' locations, in addition to the maze cleaning status in 2D format. At each step, the team gets a positive reward for each spot cleaned, and a negative reward as time cost.
    
\end{itemize}

\begin{figure}[h]
     \centering
     \subfloat[Fleet Coordination\label{Benchmarks_Fleet}]{
     \includegraphics[width=0.48\columnwidth]{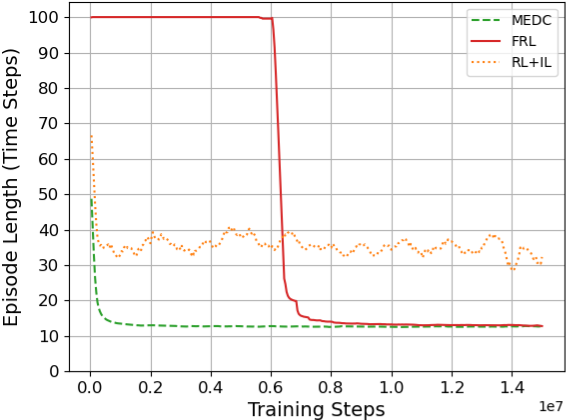}}
     \hfill
     \subfloat[Maze Cleaning\label{Benchmarks_Cleaning}]{
     \includegraphics[width=0.48\columnwidth]{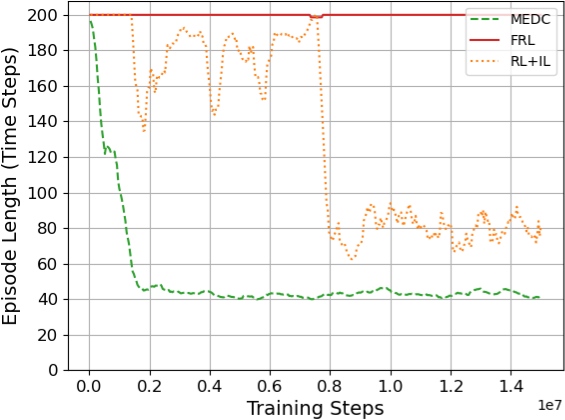}}
     \hfill
        \caption{The episodic length throughout the learning for the (a) fleet coordination and (b) maze cleaning environments.}
        \label{Results: Benchmarks2}
\end{figure}

\begin{figure}[h]
     \centering
     \subfloat[Fleet Coordination\label{BenchmarksFaulty_Fleet}]{
     \includegraphics[width=0.48\columnwidth]{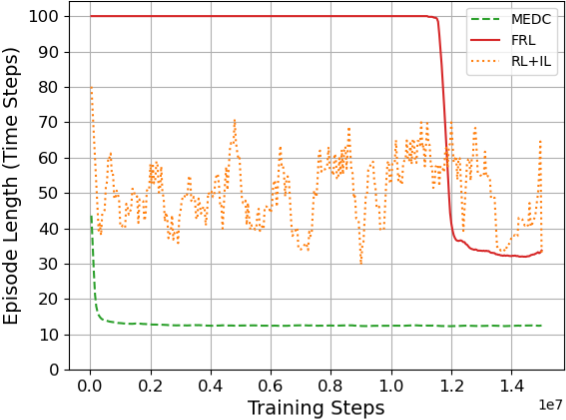}}
     \hfill
     \subfloat[Maze Cleaning\label{BenchmarksFaulty_Cleaning}]{
     \includegraphics[width=0.48\columnwidth]{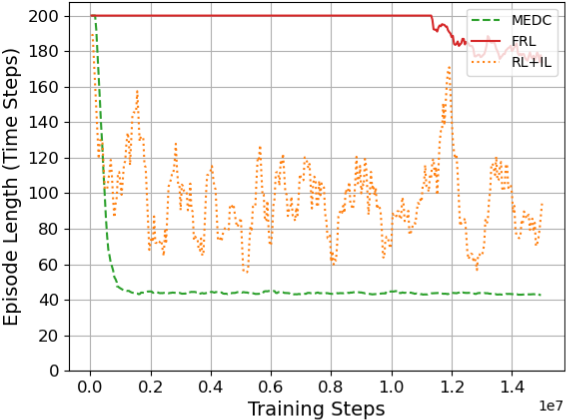}}
     \hfill
        \caption{The episodic length throughout the learning for the (a) fleet coordination and (b) maze cleaning environments, while using 3 faulty experts.}
        \label{Results: Benchmarks2Faulty}
\end{figure}

The two environments are simulated using MEDC against the FRL and IL-assisted RL benchmarks. Both environments are simulated for a 10$\times$10 grid and 3 agents, with the fleet coordination environment having 8 costumers. The episode length is used as a metric to assess the learning process, since both applications require the tasks to be finished as fast as possible. Similar to process describe in Section \ref{MEDC vs Benchmarks}, MEDC and IL-assisted RL use 6 previously trained single- and multi-agent experts from environments of varying complexities (i.e. different number of agents and costumers, and different maze complexities) to assess in training. For FRL, 7 users of varying environments are used in each application, where users share their models to update a global model that is then returned to the users. We report the performance from the perspective of a single user with an environment of 3 agents. Fig. \ref{Results: Benchmarks2} and Fig. \ref{Results: Benchmarks2Faulty} compare MEDC with benchmarks for the cases of truthful and faulty experts, respectively. Similarly to the results obtained for the target localization environment, it can be seen that MEDC outperforms the benchmarks for both environments, and under the two scenarios of truthful and faulty experts. MEDC achieves faster convergence and better performance, in terms of episodic length, even with the existence of faulty and malicious experts, showing the resiliency of the proposed method.

\subsection{Smart Contracts Complexity Analysis}
This section analyzes the complexity of the developed smart contracts in terms of gas cost. Since a Consortium Blockchain is proposed, the deployment and execution of the smart contracts do not require any payments by the users. However, the gas cost is a good measure of the complexity of the smart contracts to indicate their feasibility. Table \ref{Table: gas} presents the gas cost of the deployment and execution of the smart contracts and their functions. As seen in the table, the gas costs are low, reflecting the feasibility of the proposed contracts. For reference, we present a benchmark of the gas cost of deploying and executing a similar UMC smart contract as discussed in \cite{kadadha2022chain}.

\begin{table}[ht]
\caption{Blockchain gas cost.}
\setlength{\tabcolsep}{0pt}
\begin{center}
\begin{tabular}{|P{0.25\columnwidth}|P{0.25\columnwidth}|P{0.25\columnwidth}|}
\hline
\textbf{Contract} & \textbf{Function} & \textbf{gas cost}\\ 
\hline
\multirow{4}{*}{UMC} & deployment & 433933 \\
\cline{2-3}
& \textit{addUser()} & 35857 \\
\cline{2-3}
& \textit{updateInfo()} & 67582 \\
\cline{2-3}
& \textit{updateReputation()} & 77286 \\
\hline
\multirow{4}{*}{MMC} & deployment & 1557311 \\
\cline{2-3}
& \textit{addModel()} & 287448 \\
\cline{2-3}
& \textit{allocateModels()} & 43842 \\
\cline{2-3}
& \textit{updateModelRep()} & 79374 \\
\hline
\multirow{2}{*}{UMC - Benchmark} & deployment & 1228566\\
\cline{2-3}
& \textit{addUser()} & 352352\\
\hline
\end{tabular}

\end{center}
\label{Table: gas}
\end{table}

\section{Conclusion}
\label{Conclusion}
In this paper, the problems of sample efficiency and reward sparsity in Multi-Agent Deep Reinforcement Learning systems is tackled. A novel Blockchain-assisted Multi-Expert Demonstration Cloning (MEDC) framework is proposed, in which users share trained models to be used as expert models by other users in the MEDC method. The proposed MEDC method utilizes expert models to suggest actions to new MDRL agents, aiming to provide better experiences where the sparse reward is more frequently obtained, which speeds up the learning. Unlike methods such as FRL, the proposed MEDC method is more resilient to faulty and malicious shared models, and allows for models of different architectures to be used together. On a Consortium Blockchain, smart contracts are designed to manage the model sharing and allocation process using a Greedy method, in which attributes about the model and the users are used in the assignment process. Experiments in the target localization environment show that the proposed MEDC method speeds up the learning noticeably when compared to models trained only with sparse rewards. Additionally, it was shown that the MEDC is resilient to faulty and malicious expert models, where the performance of MEDC could be the same as MDRL with sparse reward at worse. When compared to FRL, RL with shaped rewards, and IL-assisted RL, MEDC showed dominance in terms of utilizing experiences from different environments in guiding the learning, and in being resilient against bad expert models. Moreover, the adaptability of the proposed methods is tested on two other multi-agent environments, namely Fleet Coordination and Maze Cleaning, showing dominance in learning convergence and resiliency to faulty models, when compared to FRL and IL-assisted RL. Finally, the developed smart contracts that manage users' and models' allocations are analyzed in terms of gas cost and compared to benchmarks, showing low cost that reflects their feasibility.

\bibliography{bibliography}

\begin{thebibliography}{10}
\providecommand{\url}[1]{#1}
\csname url@samestyle\endcsname
\providecommand{\newblock}{\relax}
\providecommand{\bibinfo}[2]{#2}
\providecommand{\BIBentrySTDinterwordspacing}{\spaceskip=0pt\relax}
\providecommand{\BIBentryALTinterwordstretchfactor}{4}
\providecommand{\BIBentryALTinterwordspacing}{\spaceskip=\fontdimen2\font plus
\BIBentryALTinterwordstretchfactor\fontdimen3\font minus \fontdimen4\font\relax}
\providecommand{\BIBforeignlanguage}[2]{{%
\expandafter\ifx\csname l@#1\endcsname\relax
\typeout{** WARNING: IEEEtran.bst: No hyphenation pattern has been}%
\typeout{** loaded for the language `#1'. Using the pattern for}%
\typeout{** the default language instead.}%
\else
\language=\csname l@#1\endcsname
\fi
#2}}
\providecommand{\BIBdecl}{\relax}
\BIBdecl

\bibitem{zhang2022reinforcement}
R.~Zhang, Q.~Lv, J.~Li, J.~Bao, T.~Liu, and S.~Liu, ``A reinforcement learning method for human-robot collaboration in assembly tasks,'' \emph{Robotics and Computer-Integrated Manufacturing}, vol.~73, p. 102227, 2022.

\bibitem{shurrab2022iot}
M.~Shurrab, S.~Singh, R.~Mizouni, and H.~Otrok, ``{IoT} sensor selection for target localization: A reinforcement learning based approach,'' \emph{Ad Hoc Networks}, vol. 134, p. 102927, 2022.

\bibitem{abououf2022self}
M.~Abououf, R.~Mizouni, S.~Singh, H.~Otrok, and E.~Damiani, ``Self-supervised online and light-weight anomaly and event detection for iot devices,'' \emph{IEEE Internet of Things Journal}, 2022.

\bibitem{silver2017mastering}
D.~Silver, J.~Schrittwieser, K.~Simonyan, I.~Antonoglou, A.~Huang, A.~Guez, T.~Hubert, L.~Baker, M.~Lai, A.~Bolton \emph{et~al.}, ``Mastering the game of go without human knowledge,'' \emph{nature}, vol. 550, no. 7676, pp. 354--359, 2017.

\bibitem{berner2019dota}
C.~Berner \emph{et~al.}, ``Dota 2 with large scale deep reinforcement learning,'' \emph{arXiv preprint arXiv:1912.06680}, 2019.

\bibitem{antonio2022multi}
G.-P. Antonio and C.~Maria-Dolores, ``Multi-agent deep reinforcement learning to manage connected autonomous vehicles at tomorrow's intersections,'' \emph{IEEE Transactions on Vehicular Technology}, vol.~71, no.~7, pp. 7033--7043, 2022.

\bibitem{alagha2022target}
A.~Alagha, S.~Singh, R.~Mizouni, J.~Bentahar, and H.~Otrok, ``Target localization using multi-agent deep reinforcement learning with proximal policy optimization,'' \emph{Future Generation Computer Systems}, vol. 136, pp. 342--357, 2022.

\bibitem{fan2021fault}
X.~Fan, Y.~Ma, Z.~Dai, W.~Jing, C.~Tan, and B.~K.~H. Low, ``Fault-tolerant federated reinforcement learning with theoretical guarantee,'' \emph{Advances in Neural Information Processing Systems}, vol.~34, pp. 1007--1021, 2021.

\bibitem{tianqing2021resource}
Z.~Tianqing, W.~Zhou, D.~Ye, Z.~Cheng, and J.~Li, ``Resource allocation in iot edge computing via concurrent federated reinforcement learning,'' \emph{IEEE Internet of Things Journal}, vol.~9, no.~2, pp. 1414--1426, 2021.

\bibitem{nadiger2019federated}
C.~Nadiger, A.~Kumar, and S.~Abdelhak, ``Federated reinforcement learning for fast personalization,'' in \emph{2019 IEEE Second International Conference on Artificial Intelligence and Knowledge Engineering (AIKE)}.\hskip 1em plus 0.5em minus 0.4em\relax IEEE, 2019, pp. 123--127.

\bibitem{nguyen2020deep}
T.~T. Nguyen \emph{et~al.}, ``Deep reinforcement learning for multiagent systems: A review of challenges, solutions, and applications,'' \emph{IEEE transactions on cybernetics}, vol.~50, no.~9, pp. 3826--3839, 2020.

\bibitem{hu2020learning}
Y.~Hu, W.~Wang, H.~Jia, Y.~Wang, Y.~Chen, J.~Hao, F.~Wu, and C.~Fan, ``Learning to utilize shaping rewards: A new approach of reward shaping,'' \emph{Advances in Neural Information Processing Systems}, vol.~33, pp. 15\,931--15\,941, 2020.

\bibitem{sami2022graph}
H.~Sami, J.~Bentahar, A.~Mourad, H.~Otrok, and E.~Damiani, ``Graph convolutional recurrent networks for reward shaping in reinforcement learning,'' \emph{Information Sciences}, vol. 608, pp. 63--80, 2022.

\bibitem{sami2022reward}
H.~Sami, H.~Otrok, J.~Bentahar, A.~Mourad, and E.~Damiani, ``Reward shaping using convolutional neural network,'' \emph{arXiv preprint arXiv:2210.16956}, 2022.

\bibitem{nguyen2021federated}
D.~C. Nguyen, M.~Ding, P.~N. Pathirana, A.~Seneviratne, J.~Li, and H.~V. Poor, ``Federated learning for internet of things: A comprehensive survey,'' \emph{IEEE Communications Surveys \& Tutorials}, vol.~23, no.~3, pp. 1622--1658, 2021.

\bibitem{gou2021knowledge}
J.~Gou, B.~Yu, S.~J. Maybank, and D.~Tao, ``Knowledge distillation: A survey,'' \emph{International Journal of Computer Vision}, vol. 129, pp. 1789--1819, 2021.

\bibitem{qi2021federated}
J.~Qi, Q.~Zhou, L.~Lei, and K.~Zheng, ``Federated reinforcement learning: Techniques, applications, and open challenges,'' \emph{arXiv preprint arXiv:2108.11887}, 2021.

\bibitem{nair2018overcoming}
A.~Nair \emph{et~al.}, ``Overcoming exploration in reinforcement learning with demonstrations,'' in \emph{2018 IEEE international conference on robotics and automation (ICRA)}, 2018, pp. 6292--6299.

\bibitem{vecerik2017leveraging}
M.~Vecerik, T.~Hester, J.~Scholz, F.~Wang, O.~Pietquin, B.~Piot, N.~Heess, T.~Roth{\"o}rl, T.~Lampe, and M.~Riedmiller, ``Leveraging demonstrations for deep reinforcement learning on robotics problems with sparse rewards,'' \emph{arXiv preprint arXiv:1707.08817}, 2017.

\bibitem{nguyen2021federatedd}
D.~C. Nguyen, M.~Ding, Q.-V. Pham, P.~N. Pathirana, L.~B. Le, A.~Seneviratne, J.~Li, D.~Niyato, and H.~V. Poor, ``Federated learning meets blockchain in edge computing: Opportunities and challenges,'' \emph{IEEE Internet of Things Journal}, vol.~8, no.~16, pp. 12\,806--12\,825, 2021.

\bibitem{kadadha2022context}
M.~Kadadha, S.~Singh, R.~Mizouni, and H.~Otrok, ``A context-aware blockchain-based crowdsourcing framework: Open challenges and opportunities,'' \emph{IEEE Access}, 2022.

\bibitem{zhu2021federated}
Z.~Zhu, S.~Wan, P.~Fan, and K.~B. Letaief, ``Federated multiagent actor--critic learning for age sensitive mobile-edge computing,'' \emph{IEEE Internet of Things Journal}, vol.~9, no.~2, pp. 1053--1067, 2021.

\bibitem{wang2019edge}
X.~Wang, Y.~Han, C.~Wang, Q.~Zhao, X.~Chen, and M.~Chen, ``In-edge ai: Intelligentizing mobile edge computing, caching and communication by federated learning,'' \emph{Ieee Network}, vol.~33, no.~5, pp. 156--165, 2019.

\bibitem{yu2020deep}
S.~Yu, X.~Chen, Z.~Zhou, X.~Gong, and D.~Wu, ``When deep reinforcement learning meets federated learning: Intelligent multitimescale resource management for multiaccess edge computing in 5g ultradense network,'' \emph{IEEE Internet of Things Journal}, vol.~8, no.~4, pp. 2238--2251, 2020.

\bibitem{liu2019lifelong}
B.~Liu, L.~Wang, and M.~Liu, ``Lifelong federated reinforcement learning: a learning architecture for navigation in cloud robotic systems,'' \emph{IEEE Robotics and Automation Letters}, vol.~4, no.~4, pp. 4555--4562, 2019.

\bibitem{liang2022federated}
X.~Liang, Y.~Liu, T.~Chen, M.~Liu, and Q.~Yang, ``Federated transfer reinforcement learning for autonomous driving,'' in \emph{Federated and Transfer Learning}.\hskip 1em plus 0.5em minus 0.4em\relax Springer, 2022, pp. 357--371.

\bibitem{liu2019double}
Z.~Liu and S.~Abbaszadeh, ``Double {Q}-learning for radiation source detection,'' \emph{Sensors}, vol.~19, no.~4, p. 960, 2019.

\bibitem{alagha2023multi}
A.~Alagha, R.~Mizouni, J.~Bentahar, H.~Otrok, and S.~Singh, ``Multi-agent deep reinforcement learning with demonstration cloning for target localization,'' \emph{IEEE Internet of Things Journal}, 2023.

\bibitem{baker2019emergent}
B.~Baker, I.~Kanitscheider, T.~Markov, Y.~Wu, G.~Powell, B.~McGrew, and I.~Mordatch, ``Emergent tool use from multi-agent autocurricula,'' in \emph{2020 Proc. Int. Conf. on Learning Representations (ICLR)}, 2020.

\bibitem{dong2020principled}
Y.~Dong, X.~Tang, and Y.~Yuan, ``Principled reward shaping for reinforcement learning via lyapunov stability theory,'' \emph{Neurocomputing}, vol. 393, pp. 83--90, 2020.

\bibitem{sartoretti2019primal}
G.~Sartoretti, J.~Kerr, Y.~Shi, G.~Wagner, T.~S. Kumar, S.~Koenig, and H.~Choset, ``Primal: Pathfinding via reinforcement and imitation multi-agent learning,'' \emph{IEEE Robotics and Automation Letters}, vol.~4, no.~3, pp. 2378--2385, 2019.

\bibitem{damani2021primal}
M.~Damani, Z.~Luo, E.~Wenzel, and G.~Sartoretti, ``Primal $ \_2 $: Pathfinding via reinforcement and imitation multi-agent learning-lifelong,'' \emph{IEEE Robotics and Automation Letters}, vol.~6, no.~2, pp. 2666--2673, 2021.

\bibitem{gronauer2021multi}
S.~Gronauer and K.~Diepold, ``Multi-agent deep reinforcement learning: a survey,'' \emph{Artificial Intelligence Review}, pp. 1--49, 2021.

\bibitem{schulman2017proximal}
J.~Schulman \emph{et~al.}, ``Proximal policy optimization algorithms,'' \emph{arXiv preprint arXiv:1707.06347}, 2017.

\bibitem{schulman2015high}
J.~Schulman, P.~Moritz, S.~Levine, M.~Jordan, and P.~Abbeel, ``High-dimensional continuous control using generalized advantage estimation,'' in \emph{2016 Proc. Int. Conf. on Learning Representations (ICLR)}, 2016.

\bibitem{lipowski2012roulette}
A.~Lipowski and D.~Lipowska, ``Roulette-wheel selection via stochastic acceptance,'' \emph{Physica A: Statistical Mechanics and its Applications}, vol. 391, no.~6, pp. 2193--2196, 2012.

\bibitem{alagha2022influence}
A.~Alagha, S.~Singh, H.~Otrok, and R.~Mizouni, ``Influence-and interest-based worker recruitment in crowdsourcing using online social networks,'' \emph{IEEE Transactions on Network and Service Management}, 2022.

\bibitem{kadadha2022chain}
M.~Kadadha, H.~Otrok, R.~Mizouni, S.~Singh, and A.~Ouali, ``On-chain behavior prediction machine learning model for blockchain-based crowdsourcing,'' \emph{Future Generation Computer Systems}, vol. 136, pp. 170--181, 2022.

\bibitem{benet2014ipfs}
J.~Benet, ``Ipfs-content addressed, versioned, p2p file system,'' \emph{arXiv preprint arXiv:1407.3561}, 2014.

\bibitem{wang2019optimization}
Y.~Wang, Z.~Cai, Z.-H. Zhan, Y.-J. Gong, and X.~Tong, ``An optimization and auction-based incentive mechanism to maximize social welfare for mobile crowdsourcing,'' \emph{IEEE Transactions on Computational Social Systems}, vol.~6, no.~3, pp. 414--429, 2019.

\bibitem{wang2020worker}
Y.~Wang, Y.~Gao, Y.~Li, and X.~Tong, ``A worker-selection incentive mechanism for optimizing platform-centric mobile crowdsourcing systems,'' \emph{Computer Networks}, vol. 171, p. 107144, 2020.

\bibitem{ehrhart2015auction}
K.-M. Ehrhart, M.~Ott, and S.~Abele, ``Auction fever: Rising revenue in second-price auction formats,'' \emph{Games and Economic Behavior}, vol.~92, pp. 206--227, 2015.

\bibitem{xidias2016path}
E.~Xidias, P.~Zacharia, and A.~Nearchou, ``Path planning and scheduling for a fleet of autonomous vehicles,'' \emph{Robotica}, vol.~34, no.~10, pp. 2257--2273, 2016.

\bibitem{punma2018autonomous}
C.~Punma, ``Autonomous vehicle fleet coordination with deep reinforcement learning,'' 2018.

\bibitem{jiang2021multi}
S.~Jiang and C.~Amato, ``Multi-agent reinforcement learning with directed exploration and selective memory reuse,'' in \emph{Proceedings of the 36th annual ACM symposium on applied computing}, 2021, pp. 777--784.

\bibitem{alagha2019data}
A.~Alagha, S.~Singh, R.~Mizouni, A.~Ouali, and H.~Otrok, ``Data-driven dynamic active node selection for event localization in {IoT} applications-a case study of radiation localization,'' \emph{IEEE Access}, vol.~7, pp. 16\,168--16\,183, 2019.

\bibitem{SHURRAB2023100867}
M.~Shurrab, R.~Mizouni, S.~Singh, and H.~Otrok, ``Reinforcement learning framework for uav-based target localization applications,'' \emph{Internet of Things}, p. 100867, 2023.

\bibitem{hussain2023predictive}
L.~A. Hussain, S.~Singh, R.~Mizouni, H.~Otrok, and E.~Damiani, ``A predictive target tracking framework for iot using cnn--lstm,'' \emph{Internet of Things}, vol.~22, p. 100744, 2023.

\bibitem{alagha2020rfls}
A.~Alagha, S.~Singh, H.~Otrok, and R.~Mizouni, ``{RFLS}-resilient fault-proof localization system in {IoT} and crowd-based sensing applications,'' \emph{Journal of Network and Computer Applications}, vol. 170, 2020.

\bibitem{alagha2021sdrs}
A.~Alagha, R.~Mizouni, S.~Singh, H.~Otrok, and A.~Ouali, ``{SDRS}: A stable data-based recruitment system in {IoT} crowdsensing for localization tasks,'' \emph{Journal of Network and Computer Applications}, vol. 177, p. 102968, 2021.

\bibitem{yuan2022uav}
B.~Yuan, R.~He, B.~Ai, R.~Chen, G.~Wang, J.~Ding, and Z.~Zhong, ``A uav-assisted search and localization strategy in non-line-of-sight scenarios,'' \emph{IEEE Internet of Things Journal}, 2022.

\bibitem{abualola2023matching}
H.~Abualola, R.~Mizouni, H.~Otrok, S.~Singh, and H.~Barada, ``A matching game-based crowdsourcing framework for last-mile delivery: Ground-vehicles and unmanned-aerial vehicles,'' \emph{Journal of Network and Computer Applications}, vol. 213, p. 103601, 2023.

\bibitem{lecun2015lenet}
Y.~LeCun \emph{et~al.}, ``Lenet-5, convolutional neural networks,'' \emph{URL: http://yann. lecun. com/exdb/lenet}, vol.~20, no.~5, p.~14, 2015.

\end{thebibliography}
\bibliographystyle{IEEEtran}
\end{document}